\documentclass[manuscript,screen]{acmart}
\usepackage{soul}
\usepackage{listings}
\usepackage{siunitx}
\AtBeginDocument{%
  }

\acmConference[FAccT '27]{ACM Conference on Fairness, Accountability, and Transparency }{June 27--30,
  2027}{Montréal, Canada}
\acmISBN{978-1-4503-XXXX-X/2018/06}

\title[AI Content in Political Campaigns]{Measuring Generative AI Content in Political Campaigns: Evidence of Rising Adoption and Non-Disclosure}

\author{Anna-Maria Gueorguieva}
\email{agueorg@uw.edu}
\author{Sarah Tran}
\email{saraht45@uw.edu}
\author{Nic Weber}
\email{nmweber@uw.edu}
\affiliation{%
  \institution{Information School, University of Washington}
  \city{Seattle}
  \state{Washington}
  \country{USA}
}

\begin{document}

\begin{abstract}
The use of generative AI in political campaigns has drawn public concern and policy attention. Public concerns range from distrust of candidates to skepticism about the authenticity and accuracy of political advertising, including AI-generated images and video. Nearly half of U.S. states have enacted laws that prohibit, or require disclosure of, “materially deceptive” AI-generated content (AIGC) in elections to guard against these concerns. This study measures how candidates and their campaigns produce and circulate AIGC during an election cycle. We develop two measures that estimate not only whether campaigns use AIGC, but how much they produce (magnitude) and in what forms (modality). We apply these measures to a novel dataset of text and images from the campaign websites, X accounts, and emails of 12,977 candidates for U.S. political office from 2023 to 2025. We report three findings. First, detectable AIGC from campaigns increased significantly across modalities over the observation window, and campaigns that use AIGC frequently do not disclose it. Second, incumbents’ campaigns produced significantly less AIGC than challengers’ campaigns. Third, candidates in states with AIGC disclosure or prohibition laws do not produce measurably less AIGC than candidates in other states. Our method and findings have implications for transparency and trust in elections, and on state laws that regulate AI use in campaigns. Because current state laws largely target mainly “materially deceptive” content, we identify additional policy mechanisms that could increase electoral transparency and improve trust as AIGC becomes more prevalent.

\end{abstract}



\maketitle

\section{Introduction}

\begin{figure*}
    \centering
    \includegraphics[width=.95\linewidth]{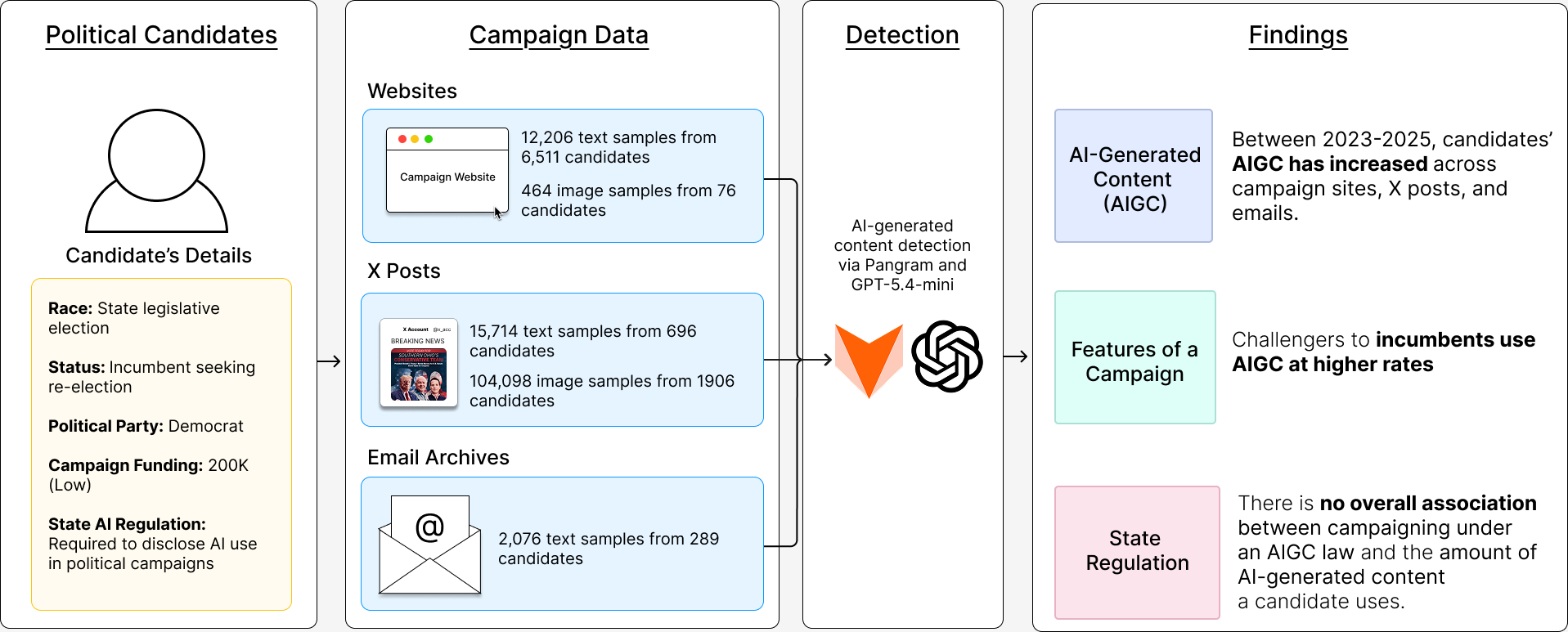}
    \caption{We collect data on political candidates across political races from 2023 to 2025 and their campaign data from three modalities (campaign websites, X posts, and emails). We apply AI detection methods to text and image content and introduce two  metrics to estimate prevalence of AIGC across races and modalities. We use these estimates to characterize detectable AIGC use over time, which types of candidates use it, and potential effects of state regulations on its use.}
    \label{fig:teaser}
\end{figure*}

Artificial intelligence (AI) tools are increasingly common across political contexts, with election campaigns using AI to conduct voter outreach \cite{haq-2024, mann-2021}, predict voter behavior \cite{tomic-2023} and produce campaign materials \cite{safiullah-2022}. The use of generative AI to create election-related content in elections is a bipartisan concern, with members of both political parties in the United States expressing worries over how AI may be used to spread misinformation and steer elections \cite{gracia-2024}, or as a tool for voter persuasion \cite{argyle2025testing}. The use of AI-generated content (AIGC) in elections by candidates can lead to distrust in democracy and elections generally, rather than just impacting one specific individual candidate \cite{jungherr-2026}. While some formal regulations have been passed at the state level, three approaches to AIGC transparency are emerging: direct disclosure, indirect disclosure, and detection. Candidates can directly disclose AI use through statements or visual captions; nearly half of U.S. states have passed laws requiring direct disclosure of "materially deceptive" AI election content. Indirect disclosures happen, for example, through platform AI content labels or campaign finance records for AI purchases. A third proactive approach is external AI detection methods, these might be used by both regulators and public interest advocates to identify undisclosed AIGC use. In the following paper, we explore the efficacy of using detection to measure detectable AI content posted by political candidates and their campaigns. We check whether that content carries the direct disclosures state laws require and whether campaigns report AI spending. To investigate detectable AI content we develop two measurement techniques to answer the following research questions:


\begin{flushleft}
\textbf{RQ 1:} What is the magnitude of detectable AI-generated content in their political campaigns? Specifically: How prevalent is it; In what modalities does it appear; and, How has the use of AIGC changed over time?
\end{flushleft}
\begin{flushleft}
\textbf{RQ 2:} Are features of a campaign (incumbency status, race type, political party, total funding, and AI in election state laws) associated with the magnitude of AI-generated content they produce?
\end{flushleft}
Beyond describing how political candidates use AIGC, a further question is whether regulation shapes that use \cite{barclay2025regulating}. Between 2019 and 2025, twenty-six U.S. states enacted laws that prohibit or require the disclosure of deceptive or harmful AIGC in elections. While these law focus on narrower, and sometimes hard to define, targets of deceptive use-cases, it is possible that just the existence of these laws may alter how candidates use AI to create campaign content. Enforcing these laws and reporting results would require independent verification of AIGC in campaign materials \cite{foos2024use}, and oversight bodies like state election boards often lack the technical capacity to monitor emerging technologies in political campaigns \cite{diakopoulos-2020, janczuk2024risks}. Whether these laws are associated with differences in candidate behavior is therefore an open empirical question:

\begin{flushleft}
\textbf{RQ 3:} What is the effect of state-level regulations on the use of AI-generated content in campaigns?
\end{flushleft}

To answer these questions, we first develop two novel measures to estimate AIGC in political messaging: 1. \textit{Magnitude} estimates the total amount of AIGC produced by political campaigns; and, 2. \textit{Modality} estimates the distribution of AIGC for each different communication channel used by a campaign. Next, we collect data on 12,977 political campaigns that took place in the USA between 2023–2025. Per candidate, our dataset contains text and image data from campaign websites, candidate emails, and social media campaign posts. We additionally collect data on features of the candidate (e.g. political party, incumbency status, funding received, and race outcomes) and election (e.g. date, outcome, partisan lean of district). We then apply our two metrics (magnitude and modality) to candidate data, and estimate the use of AIGC in political campaigns across our sample. Overall, this paper makes three research contributions: First, we provide a scalable approach to measure AIGC in campaigns and show that detectable AIGC rose significantly in U.S. elections from 2023 to 2025, though it's rarely disclosed. Second, we characterize how AIGC magnitude varies with candidate features, drawing on external data such as party and incumbency. Third, we show that candidates in states regulating AIGC use modestly less in federal races, though this association is absent in state and municipal races. We conclude by discussing how lack of transparency in AIGC may affect future efforts to ensure voter confidence in elections. We propose two policy instruments to improve transparency: collaboration to improve platform disclosures and required funding disclosures for AI spending.


\section{Related Work}

\subsection{AI content in political campaigns}
AI serves many functions in political campaigns, such as the creation of targeted advertisements, voter outreach, or to predict voter behavior \cite{martin-2024}. Both public and scholarly attention have concentrated on its politically harmful or deceptive uses.  For example, one candidate for mayor in Louisiana blamed a deepfake ad that ran against him for being one of the reasons that he lost his election \cite{swenson-2024}. While deepfakes can have an impact on elections and shape public perception, research shows their effect is more likely to erode trust in elections and media broadly rather than to move any single candidate's favorability \cite{jungherr-2026}. While deceptive uses of AI are important, especially given the erosion of public trust, scholars have begun to argue that reporting on AI in elections skews toward deepfakes and politically extreme cases, neglecting more routine uses \cite{jungherr-2026, kruschinski2026efficiency}. In this vein, Jungherr et al. make a distinction between the use of AI campaign \textit{operations} such as automated content generation and writing versus the use of AI for \textit{deceptive purposes} \cite{jungherr-2026}. Similarly, Daepp et al. argue that often AI use in elections is about generating propaganda and information dissemination at scale, rather than about deception. This can include things like ``soft-fakes" in which AI use is obvious and on purpose, or ``deep roasts" used to mock candidates. \cite{daepp-2025}. In short, the majority of previous work on AI in elections has focused specifically on deceptive deepfakes and voter trust, but has not empirically established how candidates in elections may be using generative AI to produce campaign content at scale \cite{labuz-2024, twomey-2023}.


Previous work to understand AI generated content (not just deceptive deepfakes) in elections has focused mainly on voters seeking information through chatbots \cite{jang2026inaccuracies} and the sharing of AI content via social media and television\cite{ibrahim2026impact}. On socil media, Chen et al. find that roughly 12\% of images in general-user X posts about the 2024 U.S. Presidential Election were AI-generated, and that superspreaders' AI image tweets tend to elicit more positive and less toxic responses \cite{chen-2025}. Their work is notable for its scale and its capacity to estimate AIGC volume around a specific political event using a specific medium (images). On television, Wesleyan Media Project reports that nearly \$80 million has been spent on broadcasting AI ads in the 2026 election cycle \cite{wesleyan-2026}. Other empirical work has focused on the role of platform advertising  - through Meta and Google's ad libraries - to manually look for AI use in paid advertising, ultimately creating a small-n sample due to the inability to automate searching for digitally altered materials on those platforms \cite{brennen-2025}. To summarize, prior work on AI detection in elections has focused on finding either how a general public posts on social media or how candidates pay for AI advertising through a single medium. We go beyond paid advertising and build on these studies to develop a scalable approach for estimating the total detectable AI generated content posted by candidates across mediums and platforms.



\subsection{Approaches to enhancing transparency for AI content in elections}
Given the impact that AI can have on trust in elections \cite{jungherr-2026}, we review three approaches to enhancing AIGC transparency, and by proxy trust, in elections: direct disclosures, indirect disclosures, and detection. 

\textbf{Direct disclosures} are explicitly AI-generated labels on text, images, or videos. From 2019 to 2025, twenty-four U.S. states required such labels on AI-generated content for elections, though nearly all targeted "materially deceptive" uses. These laws apply to all individuals and organizations, including the general public, campaign workers, and candidates. Most laws apply only 30 days to 120 days before an election; California's AB 2839 is the only law that regulates the use of AI post-election (see Appendix \ref{sec:state-legislations-appendix} for more information on laws and penalties). At the federal level, a recent proposal by Congresswoman Shontel Brown in October 2026 would require simulated or altered video or audio of candidates to carry either a ``AI Disclaimer" statement or a symbol with ``AI" similar to the copyright symbol \cite{brown-2026}. The current lack of litigation under these laws sets no precedent as to which types of AI uses they actually cover. Still, we might expect to see direct disclosure become more common as candidates aim to preempt enforcement or avoid public backlash by breaking evolving disclosure norms. \cite{jungherr-2026}.


 
\textbf{Indirect disclosure} occurs when a campaign's AI use is revealed through sources other than its own explicit acknowledgment, such as campaign disclosure reports. Using state-level reporting, such as Washington State's Public Disclosure Commission data\footnote{https://www.pdc.wa.gov/political-disclosure-reporting-data}, or the Federal Election Commission's campaign finance data\footnote{https://www.fec.gov/data/}, researchers can estimate campaign spexnding on AI advertisements \cite{wesleyan-2026}. These estimates are limited because finance reporting requirements are not granular: candidates may report spending on AI software together with spending on other technical or logistical tools, and candidates who use free AI tools report no related spending. Platforms that host campaign content are a second source of indirect disclosure, because they can detect and label AI use themselves. For example, Instagram automatically labels images that have been created or altered with certain AI editing tools.

\textbf{Detection} of AI is a proactive alternative when both direct and indirect disclosure are missing or fail to provide at-scale data. Detection of election content has proven successful across social media sites, especially on X to show how AI media posted by general users can spread \cite{chen-2025}. However, contemporary electioneering makes use a variety of mediums and publication channels to reach voters. The diversity of this content poses a measurement challenge for public interest advocates seeking to understand this issue \cite{diaz2026artificial}. Our work proposes two metrics that might help accurately estimate AIGC use across these mediums and channels.

\subsection{AI detection in political contexts}
AI tools let ordinary users generate text and imagery with minimal effort, opening electoral discourse to bad actors or the spreading of misinformation through new channels \cite{chang-2026, jang2026inaccuracies}. AI detection allows us to provide greater transparency on AI content in elections when direct or indirect disclosures are unavailable or not useful at scale. However, AI detection remains an open technical problem, further challenged by distributional shifts in AIGC capabilities such as evolving multimodal LLM techniques and adversarial methods for humanization \cite{ren-2026}. Conducting detection in the context of elections can be difficult beyond these challenges, as well: much political content circulates as short-form social media posts, such as TikTok videos, X posts, or Instagram photos. Lacking metadata and watermarking, such content is hard for tools to flag and demands coordinated detection across text, images, and video \cite{wirtschafter-2024}. 

Despite these many challenges, actors interested in detecting AI content - e.g. regulators, voters, and journalists - will search for tools for detection, likely turning to off-the-shelf solutions for budget and technical limitations. Thus, tool selection for AIGC in elections hinges not only empirical or theoretical performance across content types, but also on accessibility of tools. Most studies on AI detection of images or videos in political content focus on user perceptions of deepfakes \cite{momeni-2025, appel-2022} rather than actual detection of content. One study was successful in applying a multimodal LLM, GPT-4o, to detect AI use in political user generated content on X \cite{chen-2025}. This approach is well-designed for detection in political contexts as it would be accessible to lay-users, as well. 

\subsection{The importance of state and local elections}
We center our investigation of detectable AIGC in election campaigns at both the national and sub-national level in the United States. This is important for two reasons. First, public attention of AIGC in elections tends to focus on high-profile races at the national level \cite{foos2024use}. This leaves state and local elections understudied, yet prior work finds that local campaign consultants and candidates may be more likely to use generative AI than national campaigns \cite{martin-2024}. Second, the use of state and local elections allows our research to provide insight into potential impacts of state regulation that prohibits or requires the disclosure of AI use in political elections specific to a geographical place \cite{barclay2025regulating}. One reason for studying this in the U.S. context is that not all states have such laws; thus, analysis at the state level can provide insight into the effectiveness of laws when other confounding factors are fixed \cite{grumbach2022laboratories}. Furthermore, investigation of emergent technology policy shows that the regulation of AI in the United States may be most successful at the state level \cite{mallison-2024}, which allows for the incubation and spread of successful regulations as states take from each other's policies and act as ``laboratories” for policy \cite{volden-2006}. However, there is no centralized dataset of candidates across all state and relevant local elections, making it difficult to have consistent coverage; this makes it harder to study such candidates across the U.S., however, also makes it a meaningful place at which to contribute data and analysis. 

\section{Data}
We construct a dataset of 12,977 candidates' content spanning federal, state, and local elections from 2023 to 2025 (see Appendix \ref{tab:candidate-coverage} for detailed distribution by race). In the following section we describe variables we collect about each candidate's campaign.  

\subsection{Campaign level data}

\textbf{Candidate names.} We aim to obtain the names of candidates for the following local, state and federal elections: federal legislature, state legislature, mayor, governor, and state attorney general. For each race, we collect data about general-election candidates, retaining the top two by vote share when more than two candidates ran in a general election. Because no centralized dataset of U.S. political candidates exists, we combine several sources including the MIT Election Lab, Ballotpedia (a nonpartisan online political encyclopedia of American politics and elections), Wikipedia, and the National Governors Association. We obtain complete coverage of all candidates names in each race, with the exception of state legislators and U.S. House races. We manually search and verify that the missing candidates are not concentrated in any region or party.

\textbf{Campaign sites and campaign X URLs.} Using Ballotpedia, Wikipedia, and automated DuckDuckGo searches, we collect links, where possible, for each candidate's campaign website and X (social media) account. In some cases, we are collecting data for races that were completed three years in the past (2023), and candidates are no longer maintaining a web-presence. Using the Internet Archive’s Wayback Machine we were able to identify 21.7\% of all missing candidate's web presence. Where available, we also obtain campaign emails from the Archive of Political Emails (N=2,076 emails). In total, we find 4,246 X URLs and 8,171 campaign site URLs (800 of which are archived Wayback websites). We note that low coverage, roughly 2/3 of candidates we identify participating in a general election,  does not necessarily mean missing data, as some candidates may never have a campaign site or social media profile. More details about the composition of our dataset, including the number of candidate names, campaign site URLs, and X URLs, are available in Appendix Table \ref{tab:candidate-coverage}. 

\textbf{Campaign messaging content.} We collect the content of campaign websites using the Python library BeautifulSoup \footnote{\url{https://beautiful-soup-4.readthedocs.io/en/latest/}}. We collect all text from the homepage and then scrape additional text from pages using the sitemap (e.g. if the page contains a URL that appears to be about either policy (e.g. "About the issues" or "Stances") or about the candidate (e.g. "About {candidate}" or "Meet {candidate}")). Thus, we collect data from the Home page, About page, and Policy pages for each candidate, and maintain a provenance to trace where each text sample came from. For X (social media) profiles, we use Twitterapi.io \footnote{\url{https://twitterapi.io/}} to collect X posts to a linked campaign account, including the text and images, from five months before to one month after November 5th (federal election day) or on the exact date of a special election. We obtain email text from the Archive of Political Emails \footnote{\url{https://politicalemails.org/about}}. To filter the text data we use three exclusion criteria: We remove text samples under 70 tokens (the minimum required by our AI-text detector), non-English text, and exclude data from sites ending in .gov or official township websites, as we do not treat them as campaign sites. The final dataset per content type and modality is shown in Table \ref{tab:AI-labels-per-content-and-modality}.


\textbf{Additional campaign level data.} To investigate associations between campaign characteristics and detected AI content for RQ 2 we collect additional data from MIT's Election Lab, and Ballotpedia. Specifically, for each campaign we obtain: incumbency status, race type, and political party. We additionally obtain campaign funding data from OpenSecrets. For fixed effects modeling in RQ3, we also record the exact date that each election was held (which varies for special election races), and a partisan lean score for a district (covering both federal and state legislative elections) based on the outcomes of the 2024 presidential election.

\textbf{Additional campaign level data: State level regulation on AI in elections.} To analyze whether state regulation affects the prevalence of AI-generated content (RQ3), we collect state election laws on AI from 2019 to 2025 using the National Conference of State Legislatures (NCSL) database \cite{NCSL-2026}. To ensure coverage, we also conducted manual searches for AI and deepfake election laws in states not covered by NCSL. Laws are grouped by their requirements which are either 1) prohibiting deceptive AI content or 2) requiring disclosure of deceptive AI content (more information on laws in Appendix \ref{sec:state-legislations-appendix}). We consider a candidate to be subject to a given law if the law was passed and applied to the campaign cycle in which they were seeking office. We note that these laws are narrow, often regulating only ``deceptive uses". Because we are interested in potential effects on campaign behavior - rather than strict legal compliance - we still assume that pending legislation that is passed the same year as an election may still shape campaign conduct. Thus, we track whether or not a candidate was subject to prohibition or disclosure law. 

\subsection{AI detection on political content}
A central measure in our study is the amount of detectable AI-generated content across modalities. For text, we use Pangram (Version 3, the latest at the time of detection), a commercially available AI text detector. We treat its output as an estimate subject to error: detectors produce false positives, and their accuracy varies with text length, genre, and domain \cite{ren-2026}. Our choice of Pangram rests on three considerations. First, independent evaluations find that it performs well across domains and often outperforms other detectors \cite{dugan-2025, lee-2026}, with low false positive rates \cite{russell-2025-b}. Second, Pangram performs best on complete-sentence text over 100 words from domains well represented in online text. Campaign websites and emails generally meet these conditions; X posts are shorter on average, and we interpret estimates for X with corresponding caution. Third, we chose a widely available detector deliberately. Regulators, journalists, and public interest advocates who seek to monitor AI use in elections are likely to rely on tools of this kind, so our estimates indicate what such tools would report when applied to campaign communication at scale. Our aim is not to classify individual documents with certainty or to identify why campaigns use AI, but to produce consistent estimates of AIGC that support comparison across time, candidate types, and states


Because campaign web pages contain more text than other sources, we sample a contiguous 60\% of each page's body text, preserving sentence order. This provides sufficient text for detection while reducing detection costs. Following Pangram's requirements, we exclude samples shorter than 70 tokens. Mean sample lengths are 536 tokens for campaign websites, 662 for emails, and 125 for X posts. Pangram assigns each sample a label (``AI'', ``Human'', or ``Mixed'') and an AI assistance score from 0 (no AI assistance) to 1 (fully AI-generated). We classify samples labeled ``AI'' or ``Mixed'' as detected AI.
 

For image detection, we use OpenAI’s GPT-5.4-mini. Chen et al.\ \cite{chen-2025} successfully used GPT-4o in identifying AI-generated images posted by X users during the U.S. 2024 presidential election; finding a 6.5\% false positive rate and 1.6\% false negative rate over a sample of 2,000 AI labeled images for human validation. This work is the closest context to our study. Other work finds diffusion and transformer based models successful in identifying AI-generated imagery \cite{chang-2026, park-2024}. Given Chen et al.'s data similarity to ours and their effective validation, we adopt their approach to AI detection, using GPT-5.4-mini. We validate this choice of model on a sample of 340 images from Chen et al.'s study (170 AI, 170 non-AI) and find that GPT-5.4-mini agrees with their labels 78.3\% of the time, with more images being false negatives (16.5\%) than false positives (5.2\%). This is beneficial, as GPT-5.4-mini maintains a more conservative measure for labeling an image as AI. AI detection has limitations \cite{naddaf-2026} and we aim to measure \textit{detectable} AI using these tools, rather than to make generalizable or conclusive statements about all AI usage.



\section{Analysis}
Using data on detected AI use across text and image for candidates we operationalize two metrics to conduct three main analyses: 1) What is the magnitude of detectable AI-generated content in their political campaigns, 2) How are characteristics of a campaign associated with the use of AI-generated content? and 3) What is the effect of state-level regulations on the use of AI-generated content in campaigns?

\subsection{Descriptive characteristics of AIGC}
In RQ1, we seek to describe the amount of detectable AIGC across campaigns using two complementary dimensions: magnitude and modality. We interpret these as measurements of \textit{detectable} AI content, acknowledging that detection tools have imperfect accuracy. 

\textbf{Magnitude analysis.} Magnitude captures the overall prevalence of AIGC within a campaign by adding together token-length weighted text AI-assistance scores with binary AI image labels. This sum is divided by the total number of content data points (see Appendix \ref{sec:methods-appendix} for complete description of metric). Magnitude produces an outcome [0,1] with 0 representing a case where a campaign had no AI generated content, and 1 representing complete use of AI across all sampled content. Because Pangram's AI-assistance score is very rarely exactly zero, a candidate will only have an exact zero for magnitude if they only have non-AI identified images in our dataset. We report campaign AI magnitude over time and races, and apply a Mann-Whitney U test to measure if there is a significant change in magnitude across consecutive years. 


\textbf{Modality analysis.} Modality refers to the communication channel through which content is distributed (e.g., campaign site, X, email). Magnitude takes into account all content across any modality. We calculate modality using the magnitude measure for a campaign separately across campaign sites, X, and emails. We conduct an analysis aiming to understand if AIGC differs across modalities: for candidates that have magnitude scores across all three modalities, we use a Friedman test to enable a within-candidate comparison and determine whether magnitude differs significantly across modalities within the same candidate. Because we find email data on fewer candidates, we also run a two-group comparison using Mann-Whitney U test restricted to campaign sites and X, which allows us to maximize sample size for comparison of modalities within the same candidate. 



\textbf{Identifying direct disclosure.} Using our dataset of text and image content, we conduct two reviews for evidence of direct disclosure. For text, we replicate the regex patterns used by He and Bu \ \cite{he-2025} to identify AI disclosure in scientific literature. These patterns search for grammatical acknowledgments of AI. We additionally apply our own approach, filtering for common AI terminology used in research and policy. Appendix C.2 contains our regex pattern. After filtering campaign site text with the regex patterns, we conduct a manual review of any captured site text to verify the candidate's disclosure intent. For images on X, two authors conduct a manual review of 505 randomly sampled X posts that were identified to have an AI-detected image. We look for disclosure in the image itself or in the text caption posted alongside the image.

\textbf{Identifying indirect disclosure.} We then systematically review a set of campaign expense reports from the federal and state level to identify candidates spending on AI who are also in our dataset. We rely on a up-to-date dataset of candidate's that have reported AI vendors in their expense reports or described their expense as related to artificial intelligence\footnote{We use public data consolidated by N. Sanders \href{https://nsanders.me/federal-campaign-disclosure-ai/\#/}{https://nsanders.me/federal-campaign-disclosure-ai/\#/}}. This data is drawn from Federal Election Committee reports and four state level reporting agencies: Massachusetts Office of Campaign \& Political Finance, Washington Public Disclosure Commission, Colorado Secretary of State TRACER, and California Secretary of State Cal-Access. We match candidates in our dataset to the aggregated data on name, race type, and year. For names, we first match exact similar first and last name, then we match on last name and same first initial (to account for first name nicknames) if the first search couldn't find anything. We then remove candidates who are matched in both datasets but do not have any relevant content on which we run detection. We compare how many candidates in our dataset report paying for AI and the percentage of candidates with at least one AI detected content.



\subsection{Understanding variations in AIGC}
We examine patterns of association between campaign AI magnitude and features of a campaign: incumbency status, political party, total funding, race type, and if they are required to comply with a state law on AI in elections. We do not investigate demographic characteristics of the candidate as we limit features to be about a campaign, rather than a candidate; additionally, such data was not available on our original data sources. To model the relationships, we use a zero-inflated beta regression as our data contains exact zeros and many near-zeros. Our dependent variable is a candidate's AI magnitude score for a given campaign, which is a continuous variable between 0 and 1. We investigate a total of 6,285 candidates after dropping candidates with missing predictor variables (see Appendix \ref{sec:rq-2-appendix} for more information on data).

We run two beta regressions. The first uses all five independent variables we are interested in with state and year as random intercepts to account for underlying clustering. This model answers the question: controlling across race types and years, does incumbency status, race type, political party, funding for a race, or being regulated under a specific state election law have an effect on observed AI magnitude? To determine effect, we look at the comparison of average marginal effects (AMEs) between categories for the four categorical variables. For funding, a continuous variable, we look at the difference between average marginal effect on campaign AI magnitude if candidates were all at 25th percentile funding (\$19,581) versus the 75th percentile of funding (\$237,907). We further investigate three additional questions based on model interaction effects: Does incumbency effect on AI use differ across levels of office? Does fundings association to AI magnitude differ across race types? Do incumbents and challengers funding associate to AI magnitude at the same rate? Results for the second beta regression are also reported with AMEs. We use the \texttt{marginaleffects} package in R (further details on the beta regression models are in Appendix \ref{sec:rq-2-appendix} \cite{bundock-2024}.



\subsection{Impact of state regulations}
In RQ 3 we ask specifically about the impact of state-level regulations on the use of AIGC in campaigns. To answer this question, we link data about campaigns, elections, and AIGC magnitude and modality measures (from RQ 1 and RQ 2) with data about state-level regulations from NCSL (see \ref{tab:state-laws} for an overview of laws and their features). These states differ in countless fixed ways (e.g. political culture, media markets, candidate quality) that could influence both regulation and AI use by candidates. To reliably estimate a comparison between states, we use a two-way fixed effects (TWFE) linear model that absorbs both state and election-year fixed effects (a standard panel estimator for isolating within-unit, within-period variation) \cite{wooldridge2010econometric}. Our central comparison is not regulated states versus unregulated states outright, but instead we compare candidates \emph{within the same state} and \emph{within the same election year}, \textit{effectively asking whether a state's candidates used less AIGC once the state's own law was in force than comparable candidates in the same year elsewhere and in the same state before the law}. In this analysis, we assign a candidate to a ``regulated'' treatment if their election happened after a state's law took effect. We date the treatment to the day of the election rather than the calendar year, because some laws passed partway through a year. We additionally adjust for incumbency, party, campaign funding, whether the race was a special election, and the partisan lean of the district in which the election took place. Because candidates in the same state are not independent observations, we report uncertainty using standard errors that allow for correlation within states. We run this comparison separately for federal, state, and municipal offices and for all offices pooled. For significant findings we also run a battery of alternative samples and specifications reported below.

\section{Results}
\subsection{Descriptive characteristics of AIGC}

\begin{figure*}[t!]
\centering
\newcommand{\hcell}[2][c]{\renewcommand{\arraystretch}{1}\begin{tabular}[b]{@{}#1@{}}#2\end{tabular}}
\newcommand{\ncell}[2]{\renewcommand{\arraystretch}{1}\begin{tabular}[c]{@{}c@{}}#1\\[-2pt]{\scriptsize (N\,=\,#2)}\end{tabular}}

\begin{minipage}[c]{0.56\textwidth}
  \centering
  \footnotesize
  \setlength{\tabcolsep}{5pt}
  \renewcommand{\arraystretch}{2.2}
  \begin{tabular}{@{}lcccc@{}}
    \toprule
     & \multicolumn{3}{c}{\textbf{Modality}} & \\ \cmidrule(lr){2-4}
     & \textbf{Campaign site} & \textbf{X} & \textbf{Email} & \hcell{\textbf{\% detected}\\\textbf{per content}} \\
    \midrule
    \textbf{Text}  & \ncell{7.5\%}{12,206} & \ncell{8.2\%}{15,714}  & \ncell{3.7\%}{2,076} & \ncell{7.6\%}{29,996} \\
    \textbf{Image} & \ncell{10.5\%}{464}   & \ncell{6.1\%}{104,098} & --                   & \ncell{6.1\%}{104,562} \\
    \midrule
    \hcell[l]{\textbf{\% detected per}\\\textbf{modality}} & \ncell{7.6\%}{12,670} & \ncell{6.4\%}{119,812} & \ncell{3.7\%}{2,076} & \\
    \bottomrule
  \end{tabular}
\end{minipage}%
\hfill
\begin{minipage}[c]{0.40\textwidth}
  \centering
  \includegraphics[width=\linewidth]{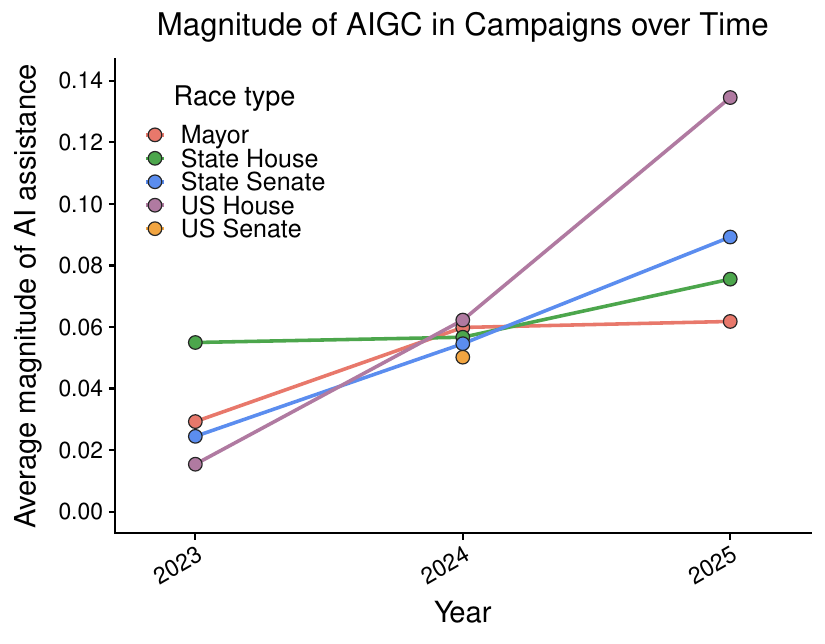}
\end{minipage}

\vspace{6pt}
\captionsetup{aboveskip=0pt, belowskip=0pt}
\begin{minipage}[t]{0.56\textwidth}
  \vspace{0pt}
  \captionof{table}{Share of content detected as AI-generated, by content type and modality.}
  \label{tab:AI-labels-per-content-and-modality}
\end{minipage}%
\hfill
\begin{minipage}[t]{0.40\textwidth}
  \vspace{0pt}
  \caption{Average magnitude of AI-generated content in campaigns over time, by race type. U.S.\ Senate elections occurred only in 2024.}
  \label{fig:overall-magnitude-over-time}
\end{minipage}
\end{figure*}


We calculate a magnitude score for all campaigns and find a consistent upward trend for the magnitude of AI assistance from 2023 to 2025, as seen in Figure \ref{fig:overall-magnitude-over-time}. A Mann Whitney U test of significance, confirms significant increase from 2024 to 2025 for State House, State Senate, and US House races (p-value $<$ 0.05 for all), though no significant increase from 2023 to 2024. For Mayoral elections, we find a significant increase from 2023 to 2024, though not from 2024 to 2025. Despite these significant changes, the magnitude of AI-generated content across campaigns is low. As Table \ref{tab:AI-labels-per-content-and-modality} shows, fewer than 10\% of all content across modality and content types is identified to be AI-generated. 




RQ1 also asks how detected AIGC differs across modalities. Our sample contains 130 candidates with magnitude scores across all three modalities (campaign sites, X, and emails). Using Friedman test, we find that \textbf{magnitude differs significantly across modalities within candidates} (Friedman chi-squared = 48.8, df = 2, p = 2.531e-11). Out of these three candidates that used all three modalities, X had the highest AI magnitude (0.071), then email (0.057), then campaign sites (0.027). Pairwise comparisons using Mann-Whitney U tests with Bonferroni correction confirm that all three modalities differ significant from one another (p < 0.0001 for all). Because we have less data from email archives, we expand our analysis to the 1,535 candidates with scores for both campaign sites and X. A Mann-Whitney U test shows the significant difference in magnitude between these channels still holds, with X content still having a greater average campaign magnitude (0.057) than campaign sites (0.047). These results suggest that candidates may use AI generated content differently across modalities, potentially with more AI content being shared on social media platforms like X. 


We reviewed direct disclosure using both regex search methods on all scraped campaign site text. We find only one mention of AI disclosure, which stated: ``this website contains content created with artificial intelligence". No further information is given to exactly which content. We do find multiple disclosure statements about using campaign website building platforms, such as Campaign Partner and Tech for Campaigns, though it's unclear if those platforms use AI. Finally, in our review of 505 AI-detected images on X posts, we find no evidence of disclosure across all posts (both by the candidate and by the platform). 

Finally, we reviewed indirect disclosures through campaign finance expense reports. In our dataset, we find 129 candidates in federal and state races (MA, WA, CO, CA) that had content available and reported spending on AI tools. Candidates most often use Prompt.io (N=18), Calltime AI (N=17), Grammarly (N=16), Otter AI (N=15), and OpenAI (N=14). Most of these platforms are for campaign outreach or logistics. Grammarly and OpenAI could be used to generate text (and one candidate pays for Anthropic). We find only a total of four candidates paying for AI that can generate image or graphical content (Descript, Midjourney, ElevenLabs, Canva AI). Out of those 129 candidates, 48\% were not identified by our detection tools to be using AI at all.




\subsection{Understanding variations in AIGC}

\begin{table}[t]
\centering
\small
\caption{Average marginal effects on campaign AI magnitude. }
\label{tab:RQ2-ame-model-1}
\begin{tabular}{lllccccc}
\toprule
Variable & Level & Reference & AME & 95\% CI & $p$ & Rel.\ change & Rel.\ 95\% CI \\
\midrule
Incumbency        & Incumbent       & Challenger       & $-0.014$ & $[-0.017, -0.010]$ & $<0.001$ & $-20.0\%$ & $[-24.6\%, -15.4\%]$ \\
                  & Open            & Challenger       & $-0.005$ & $[-0.009, -0.001]$ & $0.023$  & $-7.4\%$  & $[-13.6\%, -1.3\%]$ \\
                  & Open            & Incumbent        & $0.009$  & $[0.005, 0.012]$   & $<0.001$ & $+15.7\%$ & $[+9.0\%, +22.4\%]$ \\
\midrule
Party             & Republican      & Democrat         & $0.003$  & $[0.000, 0.006]$   & $0.028$  & $+5.3\%$  & $[+0.5\%, +10.1\%]$ \\
\midrule
Race type         & US Legislator   & State Legislator & $0.010$  & $[0.005, 0.015]$   & $<0.001$ & $+16.3\%$ & $[+7.7\%, +24.9\%]$ \\
\midrule
Compliance to law & Disclosure      & None             & $0.003$  & $[0.000, 0.007]$   & $0.044$  & $+5.7\%$  & $[+0.0\%, +11.4\%]$ \\
                  & Prohibition     & None             & $0.009$  & $[0.002, 0.015]$   & $0.010$  & $+14.8\%$ & $[+3.3\%, +26.3\%]$ \\
\midrule
Campaign funding  & 75th percentile & 25th percentile  & $0.002$  & $[0.001, 0.004]$   & $<0.001$ & $+3.8\%$  & $[+1.6\%, +6.1\%]$ \\
\bottomrule
\end{tabular}
\end{table}


Using a zero-inflated beta regression we find incumbency was the strongest predictor of AI magnitude, with challengers on average, producing more detectable AI content. On average, incumbents' campaigns had an AI magnitude 0.014 lower than challengers' (95\% CI [$-0.018$, $-0.010$]), about a 20\% less relative deduction (95\% CI [16\%, 25\%]). We find small but statistically significant differences in race type and political party, with federal races producing more detected AI generated content on average (AME = 0.01, [0.005, 0.015]) and Republicans having an average AI magnitude 0.003 greater than Democrats ([0.000, 0.006]). In terms of running in states with either a prohibition or disclosure law, average marginal effects are significantly different between candidates with disclosure laws to those without (AME = 0.003, [0.000, 0.007]) and same for prohibition (AME = 0.009, [0.002, 0.015]). Comparison of AME if candidates were at a total funding of the 25th percentile (\$19,580.73) versus 75th percentile (\$237,906.87) reveals a slight, yet significant, positive relationship (AME = 0.002 95\% CI [0.001, 0.004]). Table \ref{tab:RQ2-ame-model-1} reports the full results of average marginal effects comparison. Despite significant differences, we note that the effects are still small. The AME column in Table \ref{tab:RQ2-ame-model-1} describes the difference between two average marginal effects in the same scale as campaign magnitude (0 to 1). Thus, even though incumbency has a significant relationship with AI magnitude, incumbents are identified to have only a 0.014 difference on average, a very small change on the scale of 0 to 1.  

Using a second zero-inflated beta regression model we further explore if interactions between variables may provide more insight into which kinds of campaigns use AI. For the interaction model, non-interaction variables found in the first model still remain significant. We find no significant interaction effects when comparing the conditional AMEs between subgroups. Thus, there is no significant difference in change in AI magnitude when campaign funding increases across incumbency status; this suggests the incumbency effect likely operates through mechanisms other than solely campaign funding. Tables \ref{tab:appendix-conditional-AME-interactions}, \ref{tab:appendix-interaction-1-2}, and \ref{tab:appendix-interaction-3} in the Appendix presents the average marginal effects of variables in the interaction model and interaction effect testing for differences between conditional AMEs.


\begin{figure}[!t]
    \centering
    \includegraphics[width=0.65\columnwidth]{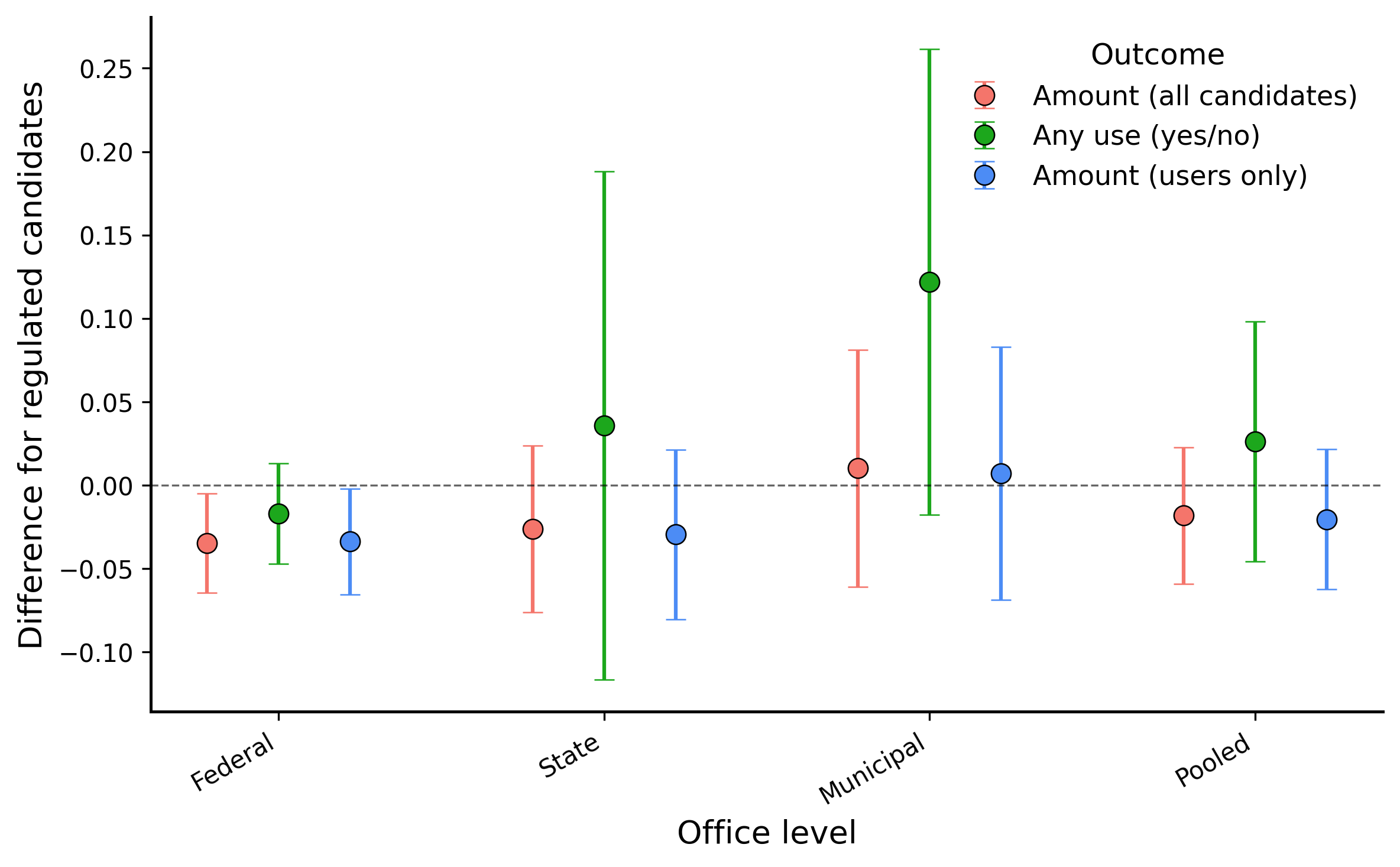}
    \caption{\textbf{Impact of state regulations at the office-level.} Each point is the estimated coefficient on the ``regulated'' indicator from a separate two-way fixed effects (TWFE) regression; The three colors portray differences in how the outcome is defined: Any use (yes/no), green = whether a candidate used any AIGC at all (binary); Amount (all candidates), red = magnitude of AIGC use across all candidates (zeros included); Amount (users only), blue = magnitude of use among only candidates who used some AIGC (zeros excluded).}
    \label{fig:rq3_coefficients}
\end{figure}

\subsection{Impact of state regulations}
To answer RQ 3, we pool all office-level candidates and find \textbf{no overall association} between campaigning under an AIGC law and the amount of AI-generated content a
candidate uses: the estimated difference is essentially zero ($-0.018$ in
transformed units; 95\% confidence interval $[-0.059, 0.023]$). However, as shown in (Table~\ref{tab:main}, Figure 3) \textbf{federal}
candidates running under a state AIGC regulation produce measurably
\emph{less} AI-generated content - a difference of $-0.035$ (95\% CI
$[-0.064, -0.005]$). The association is essentially identical when we narrow the sample to only those federal candidates who already used some AIGC, asking whether regulation is linked to how much they use rather than whether they use it at all ($-0.034$, 95\% CI [$-0.065$, $-0.002$]). By contrast, \textbf{state} and \textbf{municipal} candidates show no detectable difference (state: $-0.026$, CI $[-0.076, 0.024]$; municipal: $+0.010$, CI $[-0.061, 0.081]$ (note: the municipal sample includes only 144 candidates).
In short, candidates in regulated state-cycles use modestly less AI-generated content, but this association is confined to federal candidates and is absent at the state and municipal levels, as well as a pooled sample. Because nearly all measurable candidates use at least some AIGC, we believe that the differences identified are better explained in terms of magnitude, or \emph{how much} AI content candidates produce, not \emph{whether} they use it all. We caution these are descriptive differences among candidates whose AI use is measurable, and they are not evidence that the laws caused candidates to change their behavior. We believe these effects are useful as a baseline to measure future elections and the detectable content they generate under different regulation regimes. (full results including robustness checks for this analysis are described in Appendix \ref{sec:methods-appendix}).  

\section{Limitations}
Our work has several limitations. First, generative AI detection methods may be unable to detect all AI involvement, especially in the cases when AI was used in editing text or visual data. We took steps to mitigate this through using a detection method validated on comparable data and political contexts. We also emphasize that a primary contribution of this work is a systematic approach and metrics for estimating AI use at scale in campaigns using widely accessible detection tools; as detection methods improve future work can apply our approach to continuously improve understanding of AI-generated content in elections. Second, we retrospectively collect data on completed campaigns. In attempt to mitigate missing data we use the Wayback machine, which may introduce bias of sites that were not archived. Third, our data reflects natural imbalances as elections differ across years and states, and grouped analyses of race types may obscure meaningful variation across races. Fourth, our review of disclosure of AI-generated content is conducted only on visible campaign text which may miss parts of the website rendered dynamically. Additionally, our human review of disclosure in AI-generated content is limited to 505 photos. Additional sampling may reveal new examples or conclusions and future work should investigate the actual substance of the content. Finally, we do not analyze video content and our samples are drawn only from campaign sites, X, and email. Given the scale of our data collection and computational demands for detection, we leave video analysis to future work. Excluding videos may cause us to miss deceptive usecases that may be specific to video, including deepfake audio. Future work applying our approach and metrics to videos or additional modalities will yield valuable new information on the magnitude of AIGC in that context. 






\begin{table}[t]
\centering
\small
\caption{A summary of our primary research questions, findings, and implications.}
\label{tab:implications}
\renewcommand{\arraystretch}{1.4}
\setlength{\aboverulesep}{0pt}
\setlength{\belowrulesep}{0pt}
\newlength{\rqcol}
\setlength{\rqcol}{\dimexpr(\linewidth - 4\tabcolsep - 2\arrayrulewidth)/3\relax}
\begin{tabular}{@{}p{\rqcol}|p{\rqcol}|p{\rqcol}@{}}
\toprule
\textbf{Research question} & \textbf{Finding} & \textbf{Implication} \tabularnewline
\midrule
\raggedright \textbf{RQ1:} What is the magnitude of detectable AI-generated content in campaigns? &
\raggedright  7.6\% of text samples (N= 29,996) and 6.1\% of image samples (N=104,562) are detected to be AI; candidates rarely disclose AI use. &
\raggedright Enforcement of disclosure and prohibition laws may be difficult given the lack of transparency and monitoring capabilities. \tabularnewline
\midrule
\raggedright \textbf{RQ2:} How are the features of a campaign associated with the use of AI-generated content? &
\raggedright Challengers have, on average, higher AI magnitude than incumbents . &
\raggedright AIGC is not evenly distributed across elections and candidates, though effects are small. \tabularnewline
\midrule
\raggedright \textbf{RQ3:} What is the effect of state-level regulations on the use of AI-generated content in campaigns? &
\raggedright Federal candidates use less AIGC when regulated; there is no effect on state or municipal races. &
\raggedright Current AIGC regulations have a narrow and uneven relationship with campaign messaging. \tabularnewline
\bottomrule
\end{tabular}
\end{table}

\section{Discussion}
Table 3 summarizes our primary research questions, findings, and implications. This section reviews the implications of our findings, and two policy instruments to further improve transparency of AI content. 

\textbf{Campaigns have low rates of detected AIGC} Across all candidates and modalities, 6.5\% of content is labeled AI-generated - this is comparable to estimates in similar contexts, such as 12\% of user posted content on X for the 2024 election was AI-generated \cite{chen-2025} and 9\% of newspaper articles \cite{russell-2025-b}. The magnitude of this usage across modalities differs significantly, showing that AIGC is not evenly distributed across communication channels, even for the same candidates. Rates of detection can be due to many causes such as: candidates using AI for non-content generating purposes, limitation of detection techniques, and difficulty obtaining data post-election. Future work might track content in real time, and investigate why candidates and their campaign teams are choosing to create AI content on certain platforms at certain moments in an election. We hope that our developed metrics for estimating content across modalities might help improve estimates about the extent to which AIGC is used in electioneering.

\textbf{Lack of direct disclosure.} We review campaign website text and a small subset of AI-detected images, finding that candidates generally do not disclose AI use. We find only one website that directly states that the website content was generated with artificial intelligence. Additionally, some candidates may be unaware of if AI is being used if third-party companies contracted to build out campaign websites. While we do not systematically evaluate legislative compliance in specific states, as this would require decision-making on the "deceptiveness" of the content, we provide preliminary evidence that candidates are not disclosing AI use in generated images and text, generally. This lack of disclosure complicates goals of fostering transparency and trust in elections; and provides an early signal to potential non-compliance to disclosure regulations. This is not a new observation, as previous AI and data privacy laws requiring disclosure show sparse compliance \cite{lucas-2024, gueorguieva-2026}. Disclosure mandates across other domains also repeatedly fail to enforce compliance and transparency \cite{benshahar-2010}. In recent experimental work, Dempsey shows that voters who were shown a disclosure of AI in campaign materials report viewing candidates as less authentic and less trustworthy \cite{dempsey2026candidate}. Thus, candidates may be incentivized to not comply with regulations even when required. We suggest ways that policymakers might strengthen transparency and compliance below.  


\textbf{Lack of indirect disclosure in campaign financing.} We found that 129 candidates for which we were able to gather content data also reported spending on AI in campaign expenditures. And, out of these 129 candidates only 54\% were identified to use AI in at least one piece of content. This mismatch in detection and indirect disclosure can occur due to: limitations of detection tools, paying for AI tools that do not produce text or image content, using free AI services, or simply not reporting itemized spending. In this discussion section, we've identified multiple gaps in gaining more transparency from candidates using detection, direct disclosure, and indirect disclosure. We propose two instruments to help improve transparency: 1) responsible disclosure collaborations between candidates, social media platforms, and government bodies in charge of enforcement and 2) reporting campaign expenditures for AI.

\subsection{Recommendations}
\textbf{Campaigns, platform, and government data collaborations.} Detecting AIGC at scale is difficult without metadata. Lacking it, our approach relies on stylistic text detection and model-based image detection. Colorado's House Bill 24-1147 begins to address this by requiring AI-generated deepfakes to carry metadata identifying the tool used and the date and time of creation. Such metadata would aid detection, but gathering it across many campaigns demands substantial technical effort from enforcement agencies and researchers, and requiring generative model providers to watermark content is likely to be difficult. Platforms already hold much of this data: Meta, for instance, requires election advertisers to disclose AI-altered content and archives it in its public Ad Library \cite{meta-2026}, and Brennen et al. note that Google and Meta both collect AI-use metadata \cite{brennen-2025}. Curating this data in a centralized repository, accessible to enforcement bodies, journalists, and researchers, would reduce reliance on voluntary disclosure and manual collection. Building such infrastructure jointly with enforcement agencies could improve both compliance monitoring and research.

\textbf{Campaign expenditures as a policy instrument.} Currently, many states require disclosure of expenditures for political advertising that uses AI. For example, the Public Disclosure Commission in the state of Washington requires that candidates and committees disclose vendors, amounts paid, and information on the target audience for expenditures using synthetic media \cite{cornfield-2024}. However, these rules apply to political advertisements, not necessarily to AI use in the production of online content, campaign operations, data analysis, or voter outreach. Furthermore, campaigns may be contracting designers and consultants that use AI, creating a downstream chain of AI use that the candidates themselves may not even be aware of. Updating campaign expenditure reporting to require itemized reports for AI spending, as well as what it was used for, can increase transparency. 



\section{Conclusion}
AI-generated content (AIGC) in candidate campaign materials has been understudied, particularly at the sub-national level in the United States. Our research develops an approach and metrics for measuring AIGC across political campaigns, finding that 7.6\% of text samples and 6.1\% of image samples are detected as AI, with almost no AI disclosure. We find candidates that are challengers have, on average, a higher magnitude of AIGC than incumbents and there is little to no effect on the magnitude of detected AI when candidates are subject to either AI disclosure or prohibition in election laws at the state level. We discuss the lack of transparency in AI usage when attempting to identify AI in elections using direct disclosures, indirect disclosure, and detection. We encourage campaign expenditures requirements as a policy instrument to enforce disclosure of AI spending and collaborations with social media platforms to share metadata that provides insight into AI-detection in order to promote transparency and trust in elections. 

\newpage

\section*{Generative AI Usage Statement}
Claude Sonnet 5 and Opus 5.5 was used to assist in writing code for API calls and automating our Selenium drivers for crawling candidate pages, and improving consistency in plotting and tables. Claude Opus 5.5 was used to review R code for analyses, though all final code was written by authors. Claude Opus 5.5 was used for editing grammar and refining the clarity of sentences.

\section*{Ethical Considerations Statement}
Our study uses publicly available data and did not consist of any human subject interactions. Though we detect AIGC use through our work, we do not attribute these in any identifying way to candidates, campaigns, or political parties other than estimating the pooled sample use of AIGC by candidate affiliation. 



\clearpage
\bibliographystyle{ACM-Reference-Format}
\bibliography{sample-base}

\clearpage
\appendix

\section{State level regulation on AI in elections}
\label{sec:state-legislations-appendix}
Table \ref{tab:state-laws} presents our collected data on state laws related to the use of AI in elections. We note that more laws have been passed and will be enacted, however, our scope was laws that were enacted on or before 2025. In this table, we include a column ``explicitly only about deception". We interpret laws that are ``Yes" under this category to explicitly regulate \textit{deceptive} use cases of AI, in which AIGC can, for example, produce a ``materially different understanding or impression than a reasonable person would have from the unaltered, original version of the image, audio recording or video recording" (OR SB 1571). However, not all laws focus on deception narrowly. Other laws, such as Wisconsin focus on regulating if the AIGC was a ``means a communication that provides information about political or social issues and is made to influence the outcome of an election." (WI A 664), which is not explicitly about deception. While the New York law is specifically about deception in elections, the state also has passed the     ``artificial intelligence deceptive practices act" which states that ``a person, firm or corporation that uses for advertising purposes, or for the purposes of trade, the name, portrait or , picture, likeness, or voice of any living person without having first obtained the written consent of such person, or if a minor of his or her such minor's parent or guardian, is guilty of a misdemeanor." Thus, while the elections act itself prohibits deception, candidates may also seek relief through other acts if they ere to be impersonated with AI. 

\begin{table*}[t!]
\resizebox{\textwidth}{!}{%
\begin{tabular}{@{}lllllll@{}}
\toprule
\textbf{State} &
  \textbf{\begin{tabular}[c]{@{}l@{}}Date \\ Enacted\end{tabular}} &
  \textbf{\begin{tabular}[c]{@{}l@{}}Bill \\ Number\end{tabular}} &
  \textbf{Type} &
  \textbf{\begin{tabular}[c]{@{}l@{}}Time frame\\ Requirements\end{tabular}} &
  \textbf{Penalties} &
  \textbf{\begin{tabular}[c]{@{}l@{}}Explicitly only\\ about deception?\end{tabular}} \\ \midrule
Texas &
  11/01/2019 &
  \begin{tabular}[c]{@{}l@{}}TX BS \\ 751\end{tabular} &
  Prohibition &
  30 days before &
  \begin{tabular}[c]{@{}l@{}}Class A misdemeanor. Max of 1 year in prison \\ and/or a fine of \$4,000.\end{tabular} &
  Yes \\ \hline
Michigan &
  01/01/2023 &
  \begin{tabular}[c]{@{}l@{}}MI HB \\ 5141\end{tabular} &
  Disclosure &
  90 days before &
  \begin{tabular}[c]{@{}l@{}}First violation: Max of 90 days in prison and/or  \\ a fine of \$500.  Second violations: Max \\of five years in prison and/or a fine of \$1,000.\end{tabular} &
  No \\ \hline
Washington &
  04/13/2023 &
  \begin{tabular}[c]{@{}l@{}}WA SB \\ 5152\end{tabular} &
  Disclosure &
  None &
  \begin{tabular}[c]{@{}l@{}}A candidate may seek general or special \\ damages, including attorney's fees \\ and costs.\end{tabular} &
  No \\ \hline
Minnesota &
  08/01/2023 &
  \begin{tabular}[c]{@{}l@{}}MN HB \\ 1370\end{tabular} &
  Prohibition &
  \begin{tabular}[c]{@{}l@{}}90 days before a\\  nominating \\ convention\end{tabular} &
  \begin{tabular}[c]{@{}l@{}}First violation: Max of 90 days in prison and/or a \\ fine of \$1,000.  Second violation: Max of five \\ years in prison and/or a fine of \$10,000. \\ Violations that result in bodily harm: Max\\ one year in prison and/or a fine of \$3,000.\end{tabular} &
  Yes \\ \hline
California &
  01/01/2024 &
  \begin{tabular}[c]{@{}l@{}}CA AB \\ 2839\end{tabular} &
  Disclosure &
  \begin{tabular}[c]{@{}l@{}}120 days before, \\ 60 days after\end{tabular} &
  \begin{tabular}[c]{@{}l@{}}A candidate may seek general or special \\ damages, including attorney's fees \\ and costs.\end{tabular} &
  Yes \\ \hline
Florida &
  02/09/2024 &
  \begin{tabular}[c]{@{}l@{}}FL HB \\ 919\end{tabular} &
  Disclosure &
  None &
  \begin{tabular}[c]{@{}l@{}}A person who fails to include the disclaimer \\ is guilty of misdemeanor of the first degree.\end{tabular} &
  Yes \\ \hline
New Mexico &
  03/05/2024 &
  \begin{tabular}[c]{@{}l@{}}NM HB \\ 182\end{tabular} &
  Disclosure &
  90 days before &
  \begin{tabular}[c]{@{}l@{}}A civil penalty up to \$1,000 for each violation\\  not to exceed \$20,000 total.\end{tabular} &
  Yes \\ \hline
Indiana &
  03/12/2024 &
  \begin{tabular}[c]{@{}l@{}}IN H \\ 1133\end{tabular} &
  Disclosure &
  None &
  \begin{tabular}[c]{@{}l@{}}A candidate may seek general or special \\ damages, including attorney's fees \\ and costs.\end{tabular} &
  No \\ \hline
Utah &
  03/13/2024 &
  \begin{tabular}[c]{@{}l@{}}UT S \\ 131\end{tabular} &
  Disclosure &
  None &
  \begin{tabular}[c]{@{}l@{}}A civil penalty of up to \$1,000 for \\ each violation.\end{tabular} &
  No \\ \hline
Wisconsin &
  03/21/2024 &
  \begin{tabular}[c]{@{}l@{}}WI A \\ 664\end{tabular} &
  Disclosure &
  None &
  Forfeiture up to \$1,000 for each violation. &
  No \\ \hline
Idaho &
  03/25/2024 &
  \begin{tabular}[c]{@{}l@{}}ID H \\ 664\end{tabular} &
  Disclosure &
  None &
  \begin{tabular}[c]{@{}l@{}}A candidate may seek general or special \\ damages, including attorney's fees \\ and costs.\end{tabular} &
  Yes \\ \hline
Oregon &
  03/27/2024 &
  \begin{tabular}[c]{@{}l@{}}OR S \\ 1571\end{tabular} &
  Disclosure &
  None &
  A penalty of no more than \$10,000. &
  Yes \\ \hline
New York &
  04/20/2024 &
  \begin{tabular}[c]{@{}l@{}}NY A \\ 8808\end{tabular} &
  Disclosure &
  None &
  \begin{tabular}[c]{@{}l@{}}A candidate may seek general or special \\ damages, including attorney's fees \\ and costs.\end{tabular} &
  Yes \\ \hline
Mississippi &
  04/30/2024 &
  \begin{tabular}[c]{@{}l@{}}MS SB \\ 2577\end{tabular} &
  Disclosure &
  90 days before &
  \begin{tabular}[c]{@{}l@{}}Max of five years and/or a maximum fine\\  of \$10,000.\end{tabular} &
  Yes \\ \hline
Alabama &
  05/15/2024 &
  \begin{tabular}[c]{@{}l@{}}AL HB \\ 172\end{tabular} &
  Disclosure &
  90 days before &
  \begin{tabular}[c]{@{}l@{}}First violations: Class A misdemeanor. \\ Second violations: Class D felony.\end{tabular} &
  Yes \\ \hline
Colorado &
  05/24/2024 &
  \begin{tabular}[c]{@{}l@{}}CO H \\ 1147\end{tabular} &
  Disclosure &
  90 days before &
  \begin{tabular}[c]{@{}l@{}}A civil penalty of at least 10\% of the amount paid to \\ advertise the deepfake. A civil  penalty of at least \$100 \\ per violation if there were no payments for \\ advertising or promotion.\end{tabular} &
  Yes \\ \hline
Arizona &
  05/29/2024 &
  \begin{tabular}[c]{@{}l@{}}AZ S \\ 1359\end{tabular} &
  Disclosure &
  90 days before &
  \begin{tabular}[c]{@{}l@{}}A person may be subject to a civil penalty for \\ each day they distribute the media \\ without a disclosure.\end{tabular} &
  Yes \\ \hline
Hawaii &
  07/03/2024 &
  \begin{tabular}[c]{@{}l@{}}HI S \\ 2687\end{tabular} &
  Disclosure &
  \begin{tabular}[c]{@{}l@{}}First working day of \\ February in even \\ years to next \\ general election\end{tabular} &
  \begin{tabular}[c]{@{}l@{}}A person depicted may seek general and \\ special damages. A court may issue a \\ restraining order. A civil fine may be \\ imposed of \$1,000 per day for violating\\  a court order.\end{tabular} &
  Yes \\ \hline
\begin{tabular}[c]{@{}l@{}}New \\ Hampshire\end{tabular} &
  08/02/2024 &
  \begin{tabular}[c]{@{}l@{}}NH H \\ 1596\end{tabular} &
  Disclosure &
  90 days before &
  \begin{tabular}[c]{@{}l@{}}A candidate may seek general or special \\ damages, including attorney's fees \\ and costs.\end{tabular} &
  No \\ \hline
Delaware &
  10/09/2024 &
  \begin{tabular}[c]{@{}l@{}}DE H \\ 316\end{tabular} &
  Disclosure &
  90 days before &
  \begin{tabular}[c]{@{}l@{}}A violation is a class B misdemeanor. \\ Intent to cause bodily harm: \\ class A misdemeanor\end{tabular} &
  Yes \\ \hline
Kentucky &
  03/24/2025 &
  \begin{tabular}[c]{@{}l@{}}KY SB \\ 4\end{tabular} &
  Disclosure &
  45 days before &
  A candidate may seek attorney's fees and costs. &
  No \\ \hline
\begin{tabular}[c]{@{}l@{}}South \\ Dakota\end{tabular} &
  03/25/2025 &
  \begin{tabular}[c]{@{}l@{}}SD S \\ 164\end{tabular} &
  Disclosure &
  90 days before &
  None &
  Yes \\ \hline
\begin{tabular}[c]{@{}l@{}}North \\ Dakota\end{tabular} &
  04/10/2025 &
  \begin{tabular}[c]{@{}l@{}}ND H \\ 1167\end{tabular} &
  Disclosure &
  None &
  None &
  No \\ \hline
Montana &
  05/08/2025 &
  \begin{tabular}[c]{@{}l@{}}MT SB \\ 25\end{tabular} &
  Disclosure &
  60 days before &
  \begin{tabular}[c]{@{}l@{}}Max \$10,000 to candidate or political party.\\ Second violations: Max of 6 months in jail and/or a fine \\ of \$500. Third violations: Max 2 years in prison \\ and/or a fine of \$5000.\end{tabular} &
  No \\ \hline
Nevada &
  06/03/2025 &
  \begin{tabular}[c]{@{}l@{}}NV AB \\ 73\end{tabular} &
  Disclosure &
  None &
  None &
  No \\ \hline
\begin{tabular}[c]{@{}l@{}}Rhode \\ Island\end{tabular} &
  07/02/2025 &
  \begin{tabular}[c]{@{}l@{}}RI H \\ 5872\end{tabular} &
  Disclosure &
  90 days before &
  \begin{tabular}[c]{@{}l@{}}A candidate may seek general or special \\ damages, including attorney's fees \\ and costs.\end{tabular} &
  Yes \\
\bottomrule
\end{tabular}%
}
\caption{Summary of all U.S. state regulation for AI-generated content / use in elections enacted from 2019--2025.}

\label{tab:state-laws}
\end{table*}

\section{Data}

\begin{table*}[t!]
\centering
\footnotesize
\setlength{\tabcolsep}{5pt}
\begin{tabular}{@{}lccccc@{}}
\toprule
\textbf{Race} &
  \textbf{Year} &
  \textbf{\begin{tabular}[c]{@{}c@{}}Candidates Total\end{tabular}} &
  \textbf{\begin{tabular}[c]{@{}c@{}}\% of Names Found\end{tabular}} &
  \textbf{\begin{tabular}[c]{@{}c@{}}\% Campaign Site Coverage\end{tabular}} &
  \textbf{\begin{tabular}[c]{@{}c@{}}\% Campaign X URL Coverage\end{tabular}} \\ \midrule
Attorney General & 2023 & 6   & 100.0    & 100.0 & 100.0 \\
                 & 2024 & 20  & 100.0    & 100.0 & 80.0  \\
                 & 2025 & 2   & 100.0    & 100.0 & 100.0 \\ \midrule
Governor         & 2023 & 6   & 100.0    & 100.0 & 83.3  \\
                 & 2024 & 22  & 100.0    & 100.0 & 72.7  \\
                 & 2025 & 4   & 100.0    & 100.0 & 75.0  \\ \midrule
Mayor            & 2023 & 74  & 100.0    & 79.7  & 48.6  \\
                 & 2024 & 76  & 100.0    & 82.9  & 51.3  \\
                 & 2025 & 72  & 100.0    & 83.3  & 38.9  \\ \midrule
State House      & 2023 & 814 & 100.0**  & 40.2  & 25.5  \\
                 & 2024 & —   & 86.0*    & 61.3  & 27.0  \\
                 & 2025 & 440 & 100.0**  & 67.0 & 46.2 \\ \midrule
State Senate     & 2023 & 342 & 97.1     & 51.2  & 26.2  \\
                 & 2024 & —   & 86.0*    & 60.2  & 35.4  \\
                 & 2025 & 72  & 100.0    & 85.7  & 41.4  \\ \midrule
US House         & 2023 & 6   & 100.0**  & 100.0 & 66.7  \\
                 & 2024 & 870 & 98.7     & 96.9  & 66.1  \\
                 & 2025 & 12  & 100.0    & 100.0 & 75.3  \\ \midrule
US Senate        & 2024 & 68  & 97.0     & 100.0 & 90.9  \\ \bottomrule
\end{tabular}
\caption{Coverage of Candidates by Year and Race Type. Percentage of campaign site and X coverage are calculated from the amount of candidate names we have, not from all candidates. *For 2024, we could find no breakdown of state senate and house; we found 10,098 state legislatures total out of 11,718 and thus generalize to 86\% for both senate and house. **For state legislative elections, some general election candidates may be missing and are replaced by primary candidates.}
\label{tab:candidate-coverage}
\end{table*}

\subsection{Candidate seats available}
\label{sec:obtaining-candidate-names-appendix}
We evaluate campaign websites from candidates of state and local elections from 2023 to 2025. We focus on seven races, broadly: 
\begin{itemize}
    \item federal legislature (US Senate and House of Representatives),
    \item state legislatures (state upper and lower chambers),
    \item mayoral elections,
    \item gubernatorial elections, and
    \item state attorney generals.
\end{itemize}

We investigate only candidates in the general election, and where there are more than two we take only the two candidates receiving the proportion of the most votes. We report below the amount of candidates that, in theory, should have ran for each race type. Because of limitations on data availability and quality, not all of these candidates are included in the dataset. 

\textbf{Gubernatorial.} To select gubernatorial races, we review the National Governor's Association for data on races from 2023 to 2025 \footnote{https://www.nga.org/governors/elections/}. In 2023, three states held gubernatorial elections (Kentucky, Louisiana, Mississippi). In 2024, eleven states had gubernatorial elections, as we exclude gubernatorial elections that are not from incorporated territories (American Samoa and Puerto Rico). In 2025, there were two (New Jersey and Virginia). These include both governors that are term-limited and not up for reelection as well as races that include incumbent candidates. In total, 32 gubernatorial candidates were included in our study. 

\textbf{Mayoral.} To select mayoral races, we use Ballotpedia’s coverage of mayoral battleground races, races in the top 100 largest cities and races in state capitals. This gave us a total of 37 races in 2023 \footnote{\url{https://ballotpedia.org/United_States_mayoral_elections,_2023}} after we removed three races (Jefferson City, Missouri; Pierre, South Dakota; Irving, Texas) due to not having any competitors (default winner). In 2024 there were 38 races \footnote{\url{https://ballotpedia.org/United_States_mayoral_elections,_2024}}, after we removed three races (Glendale, Arizona; Augusta, Maine; Salem, Oregon) due to not having any competitors or information about competitors on Ballotpedia (default winner). In 2025 there were 36 elections covered \footnote{\url{https://ballotpedia.org/United_States_mayoral_elections,_2025}}, after we dropped one in Puerto Rico as we focus only on incorporated states and dropped two races where there are no competitors (Boston, Massachusetts; Plano, Texas). We save two candidates per race, and where more than two candidates are mentioned we select analysis only of the two candidates receiving the most votes. In total 119 mayoral candidates were included in our study. We note that many races are non-partisan. In these cases, we attempt to find the political party of each candidate, however, where it is unclear we keep the party as ‘nonpartisan’. 

\textbf{State legislation.} To select state legislative races for 2023, we review Ballotpedia’s “State legislative elections, 2023” page \footnote{\url{https://ballotpedia.org/State_legislative_elections,_2023}}. In total, 578 legislative seats were up for election across four states with regularly scheduled elections and 53 seats across special elections. In 2024, 5,807 seats were up for election across 44 states and 52 seats across special elections \footnote{\url{https://ballotpedia.org/State_legislative_elections,_2024}}. In 2025, 180 seats were up for election across two states and 95 special elections \footnote{\url{https://ballotpedia.org/State_legislative_elections,_2025}}. In total, we look at 13,530 candidates for state legislatures. However, we note that the exact split of this legislation is sometimes unknown due to either lack of data on individual state house and state senate races or due to some states having multi-winner elections. The states that do multi-winner elections for state house or senate are: Arizona, Idaho, Kentucky, New Jersey, North Dakota, South Dakota, Vermont, Virginia, Washington, New Hampshire, West Virginia, and Vermont. The amount of seats available per race per district are noted in our data and data analysis. 

\clearpage
\textbf{Federal legislation.} For federal legislative races, no US senate seats were up for election in 2023 and 2025. In 2024, 34 senate seats were up for election \footnote{\url{https://ballotpedia.org/United_States_Senate_elections,_2024}}. This makes a total of 68 candidates for US senate studied. For the US house, there were three special elections in 2023 \footnote{\url{https://en.wikipedia.org/wiki/2023_United_States_House_of_Representatives_elections}}. In 2024, all 435 district seats were up for election as normally scheduled. In 2025, there were six special elections \footnote{\url{https://en.wikipedia.org/wiki/2025_United_States_House_of_Representatives_elections}}. This makes a total of 888 candidates for US house studied. 

\textbf{Attorney general.} In 2023, there were three attorney general races  \footnote{\url{https://ballotpedia.org/Attorney_General_elections,_2023}}. In 2024 there were ten \footnote{\url{https://ballotpedia.org/Attorney_General_elections,_2024}}. In 2025, there was one (National Association of Attorney Generals) \footnote{\url{https://www.naag.org/news-resources/research-data/attorney-general-elections/}}. In total there were 30 attorney general candidates included in our study.

\subsection{Details on data cleaning and scraping}

\textbf{URL collection.} We obtain campaign websites and X profiles for candidates by first visiting the Ballotpedia page for each candidate. If no campaign site or campaign X is linked, we we conduct a web search using Duck Duck Go to find URLs that contain the candidates name or that contain either the first or last name, as well as a keyword for elections, including: ``forsenate", ``forcongress", ``forhouse", ``forgovernor", ``formayor", ``forcouncil", ``forassembly", ``fordelegate", ``forstaterep", ``vote", ``elect", ``campaign". We then validate that the campaign site links return a valid website, and if not we search the Wayback Machine for a saved web archive of the same URL within the same year of the election. We also validate the campaign X URLs, removing links that point to one specific post or hashtag. We additionally manually review any duplicate links across candidates to identify where a campaign site URL or X was wrongly assigned to a candidate. 

\textbf{Text and image collection.} After storing campaign site URLs and X profiles, we use the python package BeautifulSoup and TwitterAPI.io to scrape campaign sites and public campaign X profiles text and images. For campaign sites, we save all visible text on the home page. The scraper then goes through all additional pages that link out from that page, identify if any of the other pages on the campaign site contain the keywords for  ``about the candidate" (``about", ``meet") and keywords for ``policy" pages (``issue", ``issues", ``policy", ``policies", ``platform", ``agenda", ``priorities", ``priority", ``positions", ``position", ``plans", ``plan", ``vision", ``values", ``focus", "``reform"). For the policy pages, as some candidates may have individual pages per policy stance, we go one page deep into every link on the policy page, for example the url would follow the pattern: ``candidate/issues/policing". Policy pages may accidently capture privacy policies and terms of service, thus for URLs that end in /policy/terms or /policy/privacy we save them and classify them as a home page. To save resources, we collect only images from candidates that have been identified to use AI in their text. For X posts, we store both the text from the X post and the images in one row of the dataset. Video files are not stored in order to save resources. 

\textbf{Total content collected.} In total from campaign websites, we obtain 12,206 text samples from 6,511 candidates, with the average sample length being 535 tokens.  Of the text samples we collect, 53.9\% are from the home page of a campaign site, 27.0\% are from policy stances or issue pages, and 19.1\% are from ``about the candidate” pages. We also sample candidates that were identified to have text AIGC and obtain 464 images from campaign sites from 76 unique candidates. Out of our total candidates for which we have campaign site text and image data, 7.5\% are subject to prohibition laws and 32.5\% are subject to disclosure laws. From X profiles, we obtain 15,714 text samples from 696 candidates, with the average sample length being 125 tokens. We obtain 104,098 image samples from 1906 candidates' via X posts. Out of our total candidates for which we have text and image data from X, 48.3\% are subject to prohibition laws and 10.9\% are subject to disclosure laws. Finally, from email archives we obtain 2,076 samples from 289 candidates, with the average token length being 662. 

\subsection{Features of candidates}
For research question two, we obtain additional data on candidates. Table \ref{tab:features-and-source-rq2} presents the definition for each of the predictor variables used to group candidates and conduct analyses and the source of the data.

\begin{table*}[]
\resizebox{\textwidth}{!}{%
\begin{tabular}{@{}lll@{}}
\toprule
\textbf{Feature of Campaign} & \textbf{Definition}                                & \textbf{Source}                                             \\ \midrule
Race type &
  \begin{tabular}[c]{@{}l@{}}The office for which a candidate is\\ running.\end{tabular} &
  \begin{tabular}[c]{@{}l@{}}Collected while conducting data collection\\ from a range of sources: Ballotpedia, Wikipedia,\\ MIT Election Lab, FollowTheMoney.org \\ (now, OpenSecrets.org)\end{tabular} \\
Election type &
  \begin{tabular}[c]{@{}l@{}}Whether or not the election is a \\ special election\end{tabular} &
  \begin{tabular}[c]{@{}l@{}}Same sources as Race Type. This data is also sourced manually \\ for races that were not available via automated data collecting\\ or external datasets, such as for U.S. Senate 2024 races.\end{tabular} \\
Incumbency status &
  \begin{tabular}[c]{@{}l@{}}Whether the candidate was an\\ incumbent (already in office), a challenger\\ to an incumbent, or it was an open race \\ with no sitting incumbents\end{tabular} &
  \begin{tabular}[c]{@{}l@{}}This data is included on Ballotpedia. For data that we collect\\ manually we also save this information.\end{tabular} \\
Funding                      & Total contributions the campaign recieved          & FollowTheMoney.org (now, OpenSecrets.org)                   \\
Race outcome &
  Whether the candidate won or lost &
  \begin{tabular}[c]{@{}l@{}}Collected while conducting data collection\\ from a range of sources: Ballotpedia, Wikipedia,\\ MIT Election Lab, FollowTheMoney.org \\ (now, OpenSecrets.org)\end{tabular} \\
Party &
  \begin{tabular}[c]{@{}l@{}}Political party of the candidate. Can be any party, not \\ necessarily Democrat or Republican, though\\ we conduct analysis only on Democrat\\ vs Republican for simplicity.\end{tabular} &
  \begin{tabular}[c]{@{}l@{}}Collected while conducting data collection\\ from a range of sources: Ballotpedia, Wikipedia,\\ MIT Election Lab, FollowTheMoney.org \\ (now, OpenSecrets.org)\end{tabular} \\
Vote percentage              & The percentage of votes received by the candidate. & Collected from Ballotpedia, Wikipedia, and MIT Election Lab \\ \bottomrule
\end{tabular}%
}
\caption{Description and sources of features of candidate campaigns used in RQ2}
\label{tab:features-and-source-rq2}
\end{table*}

\subsection{Required compliance}
In order to measure changes in AI use that may be associated with the enactment of a state law, we first determine which legislative requirement a race was subject to. For each year, we create a list of states that are subject to an already enacted law prohibiting disclosure of AI in elections. We make the same list, per each year, of states that are subjected to an already enacted law requiring the disclosure of AI in elections. Using python, we write a script that reviews each election's state location and year, marking the variable required\_compliance as either prohibition or disclosure if the state meets the state and year criteria. If the given race is not in a state that has an enacted law for that year, required\_compliance is marked as none. Scraping campaign website data does not include exact information on when text was updated or posted. Thus, we consider any text that is produced in a state that has an enacted AI law to be under the requirement of the law, even if it is not within the specific “number of days” time frame of the law. For social media data, we store data on the date that the content was posted, collecting data that covers pre- and post election, as well pre any regulatory “number of days” before election time frame. For laws we cannot find the exact date of enactment, we set the date to be 01/01/YYYY where YYYY is the year of enactment. 

\section{Methods}
\label{sec:methods-appendix}

\subsection{Calculating magnitude and modality}

Our proposed metrics for measuring the use of AIGC in campaigns includes two measures that normalize the total amount of (magnitude) and the type (modality) of content produced. Below we describe each metric and its assumptions in detail.\\ 

\noindent\textbf{Magnitude.} Let $D_{c,t} = T_{c,t} \cup P_{c,t}$ denote the set of content units produced by candidate $c$ in cycle $t$ across two content types. Let $T_{c,t} = \{t_1, \dots, t_n\}$ represent text samples from campaign websites and X, where each unit $t_i$ has token length $w_i$ and receives an AI assistance score $f(t_i) \in [0,1]$, with 1 indicating fully AI-generated text and 0 indicating no AI involvement. Let $P_{c,t} = \{p_1, \dots, p_m\}$ represent images obtained from social media, where each image $p_j$ receives a binary detection score $f(p_j) \in \{0, 1\}$. We define modality-specific magnitude as:

$$
M_{c,t}^{\text{text}} = \frac{\sum_{t_i \in T_{c,t}} w_i \cdot f(t_i)}{\sum_{t_i \in T_{c,t}} w_i} \qquad M_{c,t}^{\text{image}} = \frac{\sum_{p_j \in P_{c,t}} f(p_j)}{|P_{c,t}|}
$$

\noindent Text magnitude is weighted by token length so that substantive content contributes proportionally more than shorter statements. We report each modality-specific magnitude separately, and define a combined campaign-level magnitude as:

$$
M_{c,t}^{\text{prob}} = \frac{|T_{c,t}|}{|D_{c,t}|} \cdot M_{c,t}^{\text{text}} + \frac{|P_{c,t}|}{|D_{c,t}|} \cdot M_{c,t}^{\text{image}}
$$

\noindent where each modality's contribution is proportional to its share of total content units. Applying a classification threshold $\tau$, we also report a threshold-based proportion $M_{c,t}^{\tau}$ that counts the length-weighted share of content units scoring at or above $\tau$. Because results may be sensitive to the choice of $\tau$, we report this measure at multiple thresholds ($\tau \in \{0.5, 0.7, 0.9\}$). We additionally define a binary adoption indicator $A_{c,t}$ that flags a campaign as having adopted generative AI if at least $\kappa$ content units exceed $\tau$, where $\kappa$ is calibrated against the detector's false positive rate so that the probability of a spurious flag is held below a specified level. Because no detector operates with perfect accuracy, all measures should be interpreted as estimates of \textit{detectable} AI content, representing a lower bound on actual use.\\

\noindent\textbf{Modality.} Campaign content spans multiple communication channels $k \in \mathcal{K}$ (e.g., campaign website, X). We compute all magnitude and adoption measures separately for each channel by restricting computation to content units of type $k$, yielding channel-specific estimates $M_{c,t}^{\text{prob},(k)}$, $M_{c,t}^{\tau,(k)}$, and $A_{c,t}^{(k)}$. This allows us to characterize not only how much AI-generated content a campaign produces, but where across communication channels it concentrates its use. Table \ref{tab:avg-mag-race-per-year} shows the average magnitude per year per race.

\subsection{Searching for AI disclosure in campaign sites}
We searched for AI disclosure in campaign sites to characterize the number of candidates that disclose AI use. To do so, we use the method introduced by He and Bu \ \cite{he-2025} (in the context of scientific writing AI disclosures) to search across text from websites and emails in our corpus with the following regular expression:

\begin{lstlisting}[breaklines=true, basicstyle=\ttfamily\small]
r"\b(?:Artificial\s+(?:General\s+)?Intelligence|AGI|Frontier\s+Models?|Open[\s-]Weight\s+Models?|General[\s-]Purpose\s+AI|GPAI|Large\s+Language\s+Models?|LLMs?|AI)\b"
\end{lstlisting}

\noindent This created a total of 14 term combinations to search for within campaign text.

\begin{table}[t!]
\resizebox{\columnwidth}{!}{%
\begin{tabular}{@{}llllllll@{}}
\toprule
 &
  Mayor &
  \begin{tabular}[c]{@{}l@{}}State \\ House\end{tabular} &
  \begin{tabular}[c]{@{}l@{}}State \\ Senate\end{tabular} &
  US House &
  US Senate &
  \begin{tabular}[c]{@{}l@{}}Attorney \\ General\end{tabular} &
  Governor \\ \midrule
\textbf{2023} & 0.029 & 0.055 & 0.024 & 0.015 & NA    & 0.006 & 0.046 \\
\textbf{2024} & 0.060 & 0.057 & 0.055 & 0.062 & 0.050 & 0.057 & 0.066 \\
\textbf{2025} & 0.062 & 0.076 & 0.089 & 0.135 & NA    & 0.003 & 0.018 \\ \bottomrule
\end{tabular}%
}
\caption{Average magnitude per year and race}
\label{tab:avg-mag-race-per-year}
\end{table}

\begin{figure}[t!]
    \centering
    \includegraphics[width=0.6\columnwidth]{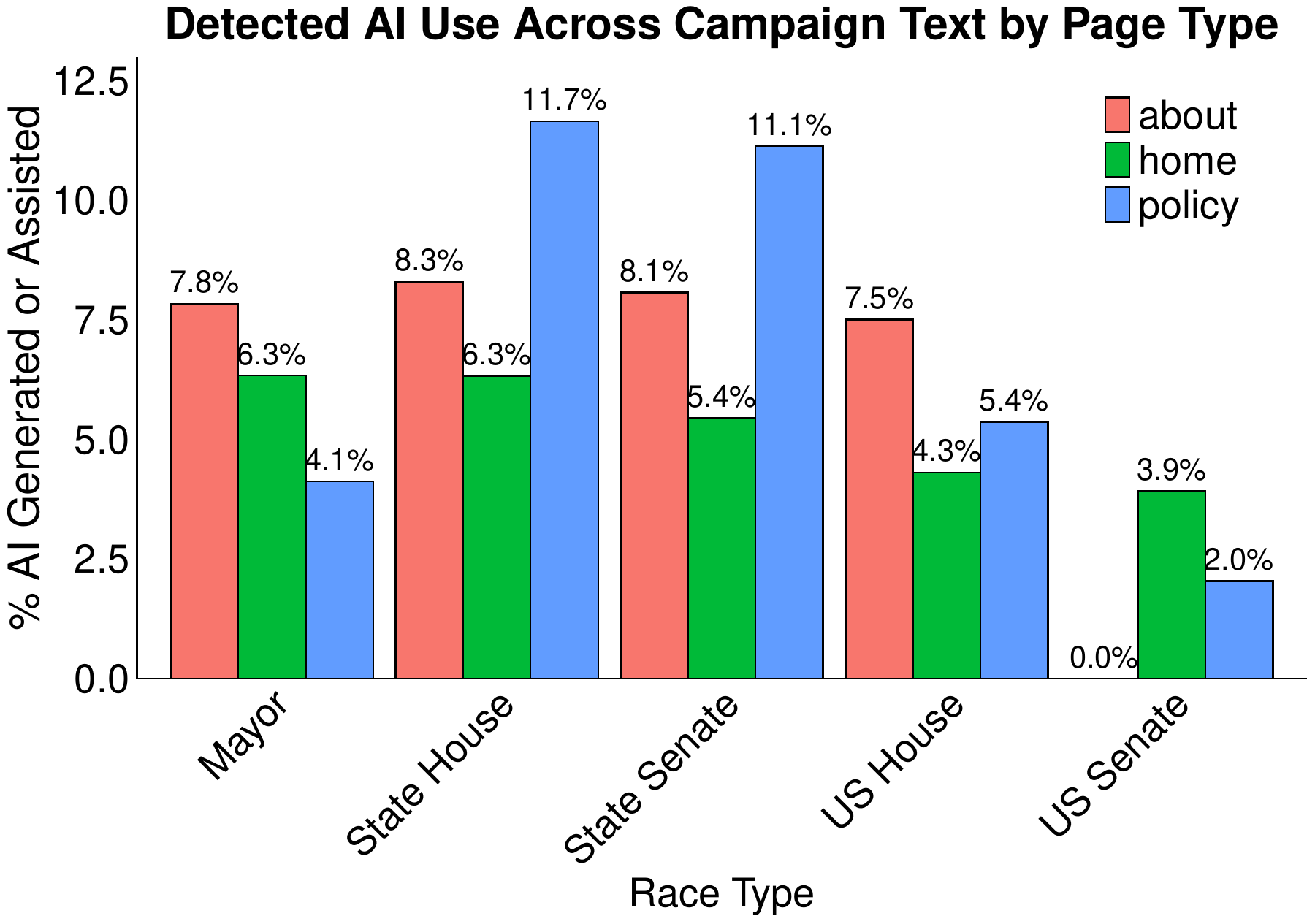}
    \caption{Percentage of text content per electoral race across years 2023-2025 identified to be AI}
    \label{fig:ai-races-campaign-text}
\end{figure}

\subsection{Beta regression for RQ 2}
\label{sec:rq-2-appendix}
This appendix specific the beta regression used to understand variations in AIGC given characteristics of a candidate. We describe the data cleaning choices and model assumptions.

\paragraph{Predictor variables:}
We begin with a data set of 7,144 candidates for which we have campaign AI magnitude calculated. To understand what features of a campaign influence AI usage (rather than post-campaign features), we exclude race outcomes (win/loss) and focus on features of political campaigns found to be important features of campaigns in political science literature: incumbency status, political party, and funding received. We also include whether or not the election occurred during a year in which a law requiring prohibition or disclosure of AI was enacted / had been already enacted. Because total funding is heavily skewed, as financial data often is, we use the log of total funding in our model. 

For the predictor variables of Party, Total Funding, and Incumbency Status, not all candidates contain this data. We drop candidates that are missing at least one of these predictor variables. For incumbency status, 5\% of candidates (N=358) do not have a known incumbency status. We drop these candidates as imputing a different value would not be a responsible data representation, and they are a small fraction of the overall data. For political parties, 48.5\% are Democrat and 48.0\% are Republican. Independents are 1.9\% and the remaining parties such as Green party or non-partisan are all less than 1\%. Because we want to look at the impact of the partisan split in America, we keep only Republicans and Democrats. Finally, 8.3\% of races are missing total funding. The races that have funding data are state legislator and US legislator races. Races without funding data are all state and local level races (mayor, attorney general, and governor). We drop these as this as funding is an important feature we want to consider in this investigation, as we hypothesize that lack of financial resources may be tied to the usage of AI. Local and state races are also more likely to have non-partisan races, causing a lack in party affiliation. In total, we drop 774 candidates that are missing any one of these three predictor variables as we do not have enough data to run the regression responsibly on the set of these 774 candidates. However, as most of these candidates are local and state candidates, this is a group that warrants future attention, especially as work on AI adoption moves further down ballot. 

We check these features for colinearity using the VIF package in R. All VIF scores are below five, identifying that we do not have the issue of colinearity in our data. 

\paragraph{Excluded data:} We exclude special elections because only 3.4\% of elections in our dataset are marked as special elections, causing a skew that may make it harder to realibility interpret results. We also exclude seats available (e.g when multiple candidates run for 2 open seats) as 88\% of our elections occur in a district with only one seat available, with the remainder of seats available (2,3 or up to 10) taking up smaller percentages that make this a less useful feature that complicates interpretation. Additionally, the number of seats available in a race is, for our analysis, not as pertinent to understand as other campaign features.

\paragraph{Random intercepts:} We set race type, state, and year as random intercepts to account for underlying clustering. Race type and state capture differences across race contexts. Year accounts for the fact that races are not evenly distributed across years. We do not consider election id to be a random intercept, as random intercepts need multiple observations per group to estimate variance and it causes the model to fail convergence. 

\paragraph{Handling missing data:} 
For the predictor variables of Party, Total Funding, and Incumbency Status, not all candidates contain this data. We drop candidates that are missing at least one of these predictor variables. For incumbency status, 5\% of candidates (N=358) do not have a known incumbency status. We drop these candidates as imputing a different value would not be a responsible data representation, and they are a small fraction of the overall data. For political parties, 48.5\% are Democrat and 48.0\% are Republican. Independents are 1.9\% and the remaining parties such as Green party or non-partisan are all less than 1\%. Because we want to look at the impact of the partisan split in America, we keep only Republicans and Democrats. Finally, 8.3\% of races are missing total funding. The races that have funding data are state legislator and US legislator races. Races without funding data are all state and local level races (mayor, attorney general, and governor). We drop these as this as funding is an important feature we want to consider in this investigation, as we hypothesize that lack of financial resources may be tied to the usage of AI. In total, we drop 859 candidates that are missing any one of these three predictor variables as we do not have enough data to run the regression responsibly on the set of these candidates. 

\paragraph{Beta regression, first model:} We use the glmmTMB model since the other R option for beta regressions, the betareg library, can’t handle random intercepts. The final model using the features describe above is:

\begin{verbatim}
glmmTMB(campaign_ai_magnitude ~ incumbency_status + party + required_compliance +
          log(total_funding + 1) + (1 | race_type) + (1 | year) + (1 | state),
        ziformula = ~1,
        data = final_set,
        family = beta_family())
\end{verbatim}

\paragraph{Beta regression, interaction model:}

\begin{verbatim}
glmmTMB(campaign_ai_magnitude ~ incumbency_status + party + required_compliance +
    log(total_funding + 1) + race_type_binary +
    log(total_funding + 1):incumbency_status +
    log(total_funding + 1):race_type_binary +
    incumbency_status:race_type_binary + (1 | year) + (1 | state) ,
  ziformula = ~1,
  data = final_set,
  family = beta_family()
)

\end{verbatim}

\begin{table}[htbp]
\centering
\newcommand{\B}[1]{{\boldmath\bfseries #1}}
\caption{Average marginal effects on campaign AI magnitude, interaction model. }
\label{tab:appendix-conditional-AME-interactions}
\resizebox{\textwidth}{!}{%
\begin{tabular}{llllccccc}
\toprule
Variable &  Level & Reference & AME & 95\% CI & $p$ & Rel.\ change & Rel.\ 95\% CI \\
\midrule
Incumbency  & Incumbent & Challenger & $-0.014$ & $[-0.018, -0.010]$ & $<0.001$ & $-20.5\%$ & $[-25.1\%, -15.8\%]$ \\
                & Open     & Challenger & $-0.006$ & $[-0.010, -0.001]$ & $0.015$  & $-8.2\%$ & $[-14.5\%, -1.9\%]$ \\
               & Open    & Incumbent  & $0.008$  & $[0.005, 0.012]$   & $<0.001$ & $+15.4\%$ & $[+8.5\%, +22.4\%]$ \\
\midrule
Party  & Republican & Democrat & $0.003$ & $[0.000, 0.006]$ & $0.028$ & $+5.3\%$ & $[+0.5\%, +10.1\%]$ \\
\midrule
Race type  & US Legislator & State Legislator & $0.008$& $[0.001, 0.014]$ & $0.024$& $+12.9\%$ & $[+1.6\%, +24.2\%]$ \\
\midrule
Compliance law  & Disclosure  & None & $0.003$ & $[0.000, 0.007]$ & $0.041$ & $+5.8\%$  & $[+0.1\%, +11.4\%]$ \\
                    & Prohibition & None & $0.009$ & $[0.002, 0.015]$ & $0.009$ & $+15.1\%$ & $[+3.6\%, +26.5\%]$ \\
\midrule
Campaign funding & 75th pct.\ (237,907) & 25th pct.\ (19,581) & $0.002$ & $[0.001, 0.004]$ & $0.002$ & $+3.8\%$ & $[+1.4\%, +6.2\%]$ \\

\bottomrule
\end{tabular}%
}
\end{table}

\begin{table}[htbp]
  \centering
  \caption{Differences in the effect of an interquartile increase in campaign funding, by incumbency status and race type. Estimates are differences in conditional AMEs when moving total funding from 25th percentile to 75th pecentile.}
  \label{tab:appendix-interaction-1-2}
  \begin{tabular}{p{3cm}lcccccc}
    \toprule
    Question & Hypothesis & Estimate & Std. Error & $z$ & $p$ & S & 95\% CI \\
    \midrule
    {oes the IQR funding effect differ by incumbency?}
     & Incumbent $-$ Challenger & $-0.000683$ & 0.00154 & $-0.442$ & 0.658 & 0.6 & [$-0.00371$, 0.00234] \\
     & Open $-$ Challenger      & $-0.002324$ & 0.00179 & $-1.302$ & 0.193 & 2.4 & [$-0.00582$, 0.00118] \\
     & Open $-$ Incumbent       & $-0.001641$ & 0.00177 & $-0.929$ & 0.353 & 1.5 & [$-0.00510$, 0.00182] \\
    \midrule
    {Does the IQR funding effect differ by race type?}
     & US Legislator $-$ State Legislator & 0.00201 & 0.00212 & 0.949 & 0.343 & 1.5 & [$-0.00214$, 0.00616] \\
    \bottomrule
  \end{tabular}

\end{table}

\begin{table}[htbp]
  \centering
  \caption{Differences in the effect of race type by incumbency status. Estimates are differences in conditional AMEs of race type (U.S. legislature vs state legislature) across incumbency groups}
  \label{tab:appendix-interaction-3}
  \begin{tabular}{p{3cm}lcccccc}
    \toprule
    Question & Hypothesis & Estimate & Std. Error & $z$ & $p$ & S & 95\% CI \\
    \midrule
    {Does the race-type effect differ by incumbency?}
     & Incumbent $-$ Challenger & $-0.00472$ & 0.00567 & $-0.832$ & 0.406 & 1.3 & [$-0.0158$, 0.00640] \\
     & Open $-$ Challenger      & $-0.00719$ & 0.00811 & $-0.886$ & 0.376 & 1.4 & [$-0.0231$, 0.00872] \\
     & Open $-$ Incumbent       & $-0.00247$ & 0.00759 & $-0.326$ & 0.745 & 0.4 & [$-0.0173$, 0.01240] \\
    \bottomrule
  \end{tabular}

\end{table}

\subsection{Fixed Effects Modeling for RQ 3}
This appendix specifies the two-way fixed effects model used to estimate the relationship between state AIGC regulation and candidate AI use, and reports the full results summarized in the main text. We first describe the analysis sample and outcome variable, then the model specification and identifying assumptions.

\paragraph{Data:}
The analysis sample consits of 7,176 candidate-races spanning 51 states (including the District of Columbia) and the 2023--2026 cycles. We group offices into three levels: federal (U.S.\ House and Senate; 817 candidates), state (governor, attorney general, state legislature, and state supreme court; 6,205), and municipal (mayor; 144). The outcome is a continuous ``AIGC amount'' score built from each candidate's tweets, email archives and campaign-website text. Because the score is skewed and bounded below by zero, we apply an inverse-hyperbolic-sine transformation - a standard log-like rescaling that compresses large values but, unlike a logarithm, is defined at zero. Coefficients are therefore in transformed units; for the small values typical here they approximate proportional differences.

\paragraph{Handling unmeasurable candidates.}
Many candidates carry a ``no\_relevant\_data'' marker, meaning no scorable
campaign content was found. We treat this conservatively: a candidate is
scored as a true zero only if we have some of their content but detected
no AI use (67 candidates); if we have no content for them at all
($\sim$5,795), we treat the score as missing and drop them, because a
blank is not the same as a measured zero. This removes roughly 45\% of
candidates, so every result below describes the subpopulation of
candidates whose AI use we can actually measure---not all candidates.

\paragraph{Reporting Comparisons}
Our within-state-and-year design uses states whose regulatory status \emph{changed} across the cycles we observe; 14 states
provide this kind of before-and-after variation, which is the minimum we
required in advance. We validated the treatment coding against an
independent compliance field stored in the data and found them to agree
on all but 0.2\% of rows. We note two important limits: First,
the district-lean control is only available for U.S.\ House and mayoral
races (where district size implies vote returns); for statewide and
legislative races the district lean is undefined. In sum, the lean we adjust for only operates in the federal and municipal models. Second, almost every
measurable candidate ($\sim$97\%) shows \emph{some} AI-generated content,
so a yes/no ``any AIGC'' version of the outcome has too little variation
to be informative; we rely instead on the magnitude (how much) outcome
and on a version restricted to candidates with positive scores (as described in detail above).

\paragraph{Estimation and robustness.}
Models were fit with the \texttt{pyfixest} package using state-clustered
standard errors. We re-ran the federal specification under 9 alternative scenarios: It is unchanged with an untransformed outcome ($-0.035$), after dropping narrowly-timed laws ($-0.034$), and in U.S.\ House races only, keeping the district-lean control where it is slightly larger ($-0.037$, 95\% CI $[-0.067, -0.008]$). Further, defining regulation from the compliance field rather than from election dates, also suggests the same direction as other controls and is larger ($-0.063$, 95\% CI $[-0.073, -0.052]$). We could not separate disclosure from prohibition laws, because only two states prohibit and neither changed status within our window.

\begin{figure}[t]
  \centering
  \includegraphics[width=\columnwidth]{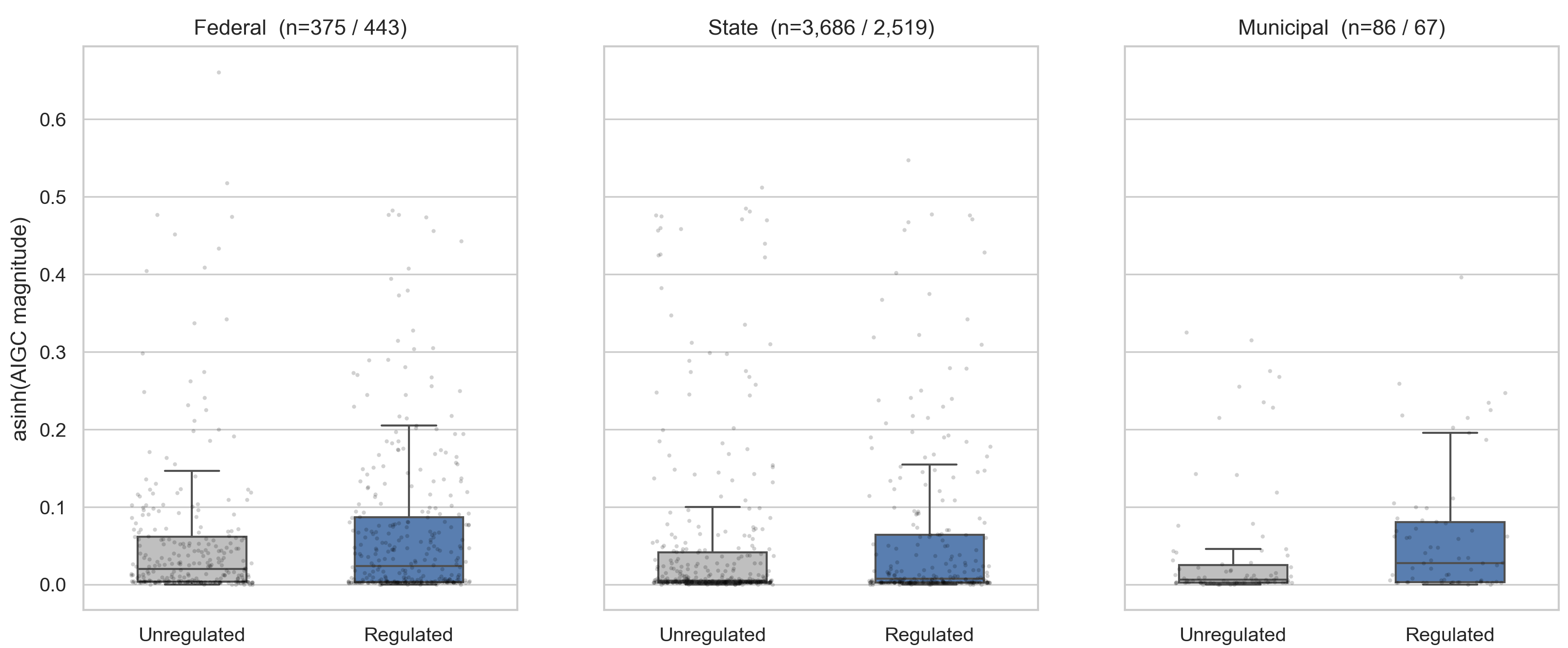}
  \caption{Raw distribution of AIGC amount (inverse-hyperbolic-sine
  scale) among candidates in unregulated versus regulated state-cycles,
  shown separately for federal, state, and municipal offices. Boxes give
  the median and interquartile range; points are individual candidates.}
  \label{fig:dist}
\end{figure}

\begin{table}[t]
  \centering
  \footnotesize
  \setlength{\tabcolsep}{4pt}
  \begin{tabular}{@{}ll S[table-format=-1.3,detect-weight=true] c r@{}}
    \toprule
    Office & Outcome & {Diff.} & {95\% CI} & {$n$} \\
    \midrule
    Federal   & Amount, all   & \bfseries -0.035 & $[-0.064,-0.005]$ & 817 \\
              & Any use       & -0.017           & $[-0.047,0.013]$  & 817 \\
              & Amount, users & \bfseries -0.034 & $[-0.065,-0.002]$ & 803 \\
    \addlinespace
    State     & Amount, all   & -0.026 & $[-0.076,0.024]$ & 6205 \\
              & Any use       &  0.036 & $[-0.117,0.188]$ & 6205 \\
              & Amount, users & -0.029 & $[-0.080,0.021]$ & 5984 \\
    \addlinespace
    Municipal & Amount, all   &  0.010 & $[-0.061,0.081]$ & 144 \\
              & Any use       &  0.122 & $[-0.018,0.262]$ & 144 \\
              & Amount, users &  0.007 & $[-0.069,0.083]$ & 136 \\
    \addlinespace
    Pooled    & Amount, all   & -0.018 & $[-0.059,0.023]$ & 7175 \\
              & Any use       &  0.026 & $[-0.046,0.098]$ & 7175 \\
              & Amount, users & -0.020 & $[-0.063,0.022]$ & 6933 \\
    \bottomrule
  \end{tabular}
  \caption{Coefficient on \emph{regulated} from two-way fixed effects
  regressions, by office level. \textbf{Bold} marks 95\% CIs excluding
  zero. Estimates are in inverse-hyperbolic-sine units (yes/no for
  ``Any use''). Outcomes: \emph{Amount, all} $=$ AIGC amount over all
  candidates (non-users $=0$); \emph{Any use} $=$ used any AIGC;
  \emph{Amount, users} $=$ amount among users only. All models absorb
  state and year fixed effects and adjust for incumbency, party, log
  funding, special-election status, and district lean; SEs clustered by
  state.}
  \label{tab:main}
\end{table}

\end{document}